\documentclass{article}

\usepackage[preprint]{neurips_2026}

\usepackage[utf8]{inputenc} 
\usepackage[T1]{fontenc}    
\usepackage{hyperref}       
\usepackage{url}            
\usepackage{booktabs}       
\usepackage{amsfonts}       
\usepackage{amsmath}
\usepackage{amssymb}
\usepackage{nicefrac}       
\usepackage{microtype}      
\usepackage{xcolor}         
\usepackage{graphicx}

\usepackage{algorithm}
\usepackage{algpseudocode}
\usepackage{mathtools}
\usepackage{multirow}
\usepackage{array}
\usepackage{enumitem}
\usepackage{caption}
\usepackage{subcaption}
\usepackage{tcolorbox}
\usepackage{xspace}
\usepackage{tabularx}
\usepackage{wrapfig}
\usepackage{pifont}
\usepackage[table]{xcolor}
\newcommand{\cmark}{\ding{51}}%
\newcommand{\xmark}{\ding{55}}%

\newcommand{\mmmc}{VisualClaw\xspace}
\newcommand{\mmbench}{VisualClawArena\xspace}

\makeatletter
\DeclareRobustCommand\onedot{\futurelet\@let@token\@onedot}
\def\@onedot{\ifx\@let@token.\else.\null\fi\xspace}

\def\eg{\emph{e.g}\onedot} 
\def\ie{\emph{i.e}\onedot}

\def\wrt{w.r.t\onedot} 

\makeatother

\title{\raisebox{-0.13\height}{\includegraphics[width=0.32\textwidth]{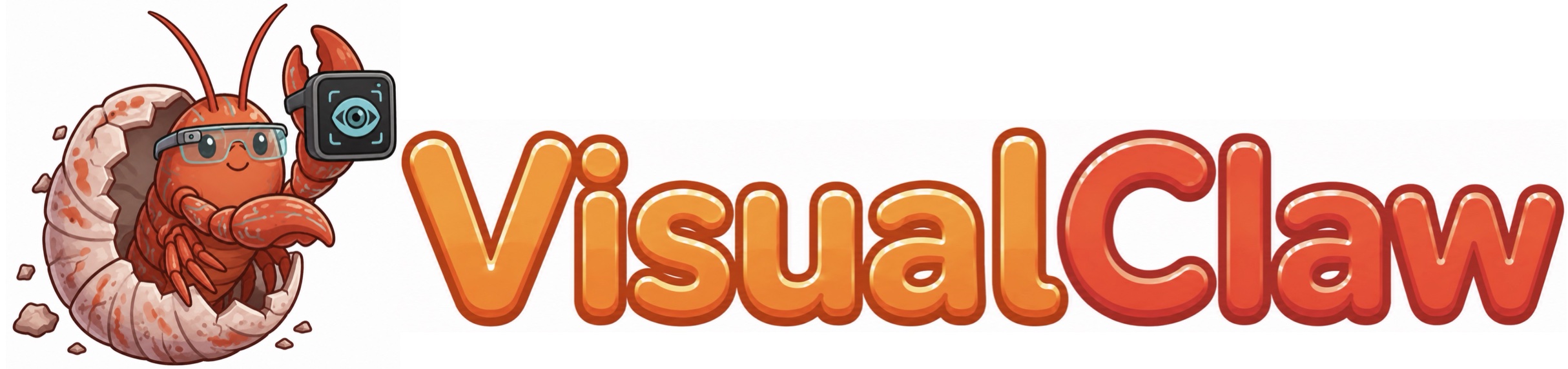}}: A \textsc{Real-Time}, \textsc{Personalized} Agent for the Physical World}

\author{%
  \textbf{Haoqin Tu$^{1*}$}\quad
  \textbf{Jianwen Chen$^{2*}$}\quad 
  \textbf{Zijun Wang$^{1}$}\quad 
  \textbf{Siwei Han$^{2}$}\quad
  \textbf{Juncheng Wu$^{1}$}\quad
  \textbf{Hardy Chen$^{1}$}\quad \vspace{.2em}\\
  \textbf{Haonian Ji$^{2}$}\quad 
  \textbf{Kaiwen Xiong$^{2}$}\quad 
  \textbf{Jiaqi Liu$^{2}$}\quad 
  \textbf{Peng Xia$^{2}$}\quad 
  \textbf{Jieru Mei$^{3}$}\quad 
  \textbf{Hongliang Fei$^{3}$}\quad \vspace{.2em}\\
  \textbf{Jason Eshraghian$^{1}$}\quad
  \textbf{Zeyu Zheng$^{4}$}\quad
  \textbf{Yuyin Zhou$^{1}$}\quad
  \textbf{Huaxiu Yao$^{2}$}\quad
  \textbf{Cihang Xie$^{1}$} \vspace{.3em}\\
  \small * equal technical contribution
  \vspace{.5em}\\
  \small
  $^{1}$UC Santa Cruz \quad $^{2}$UNC-Chapel Hill \quad $^{3}$Google \quad $^{4}$UC Berkeley \vspace{.5em}\\
\small{\hspace{3em} \includegraphics[height=1em]{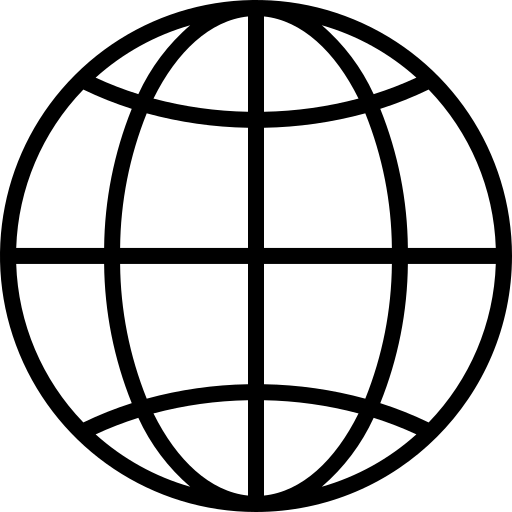} \textbf{Project Page:} \url{https://ucsc-vlaa.github.io/VisualClaw}}
\\
}

\begin{document}

\maketitle

\begin{figure*}[h]
\centering
\includegraphics[width=\textwidth]{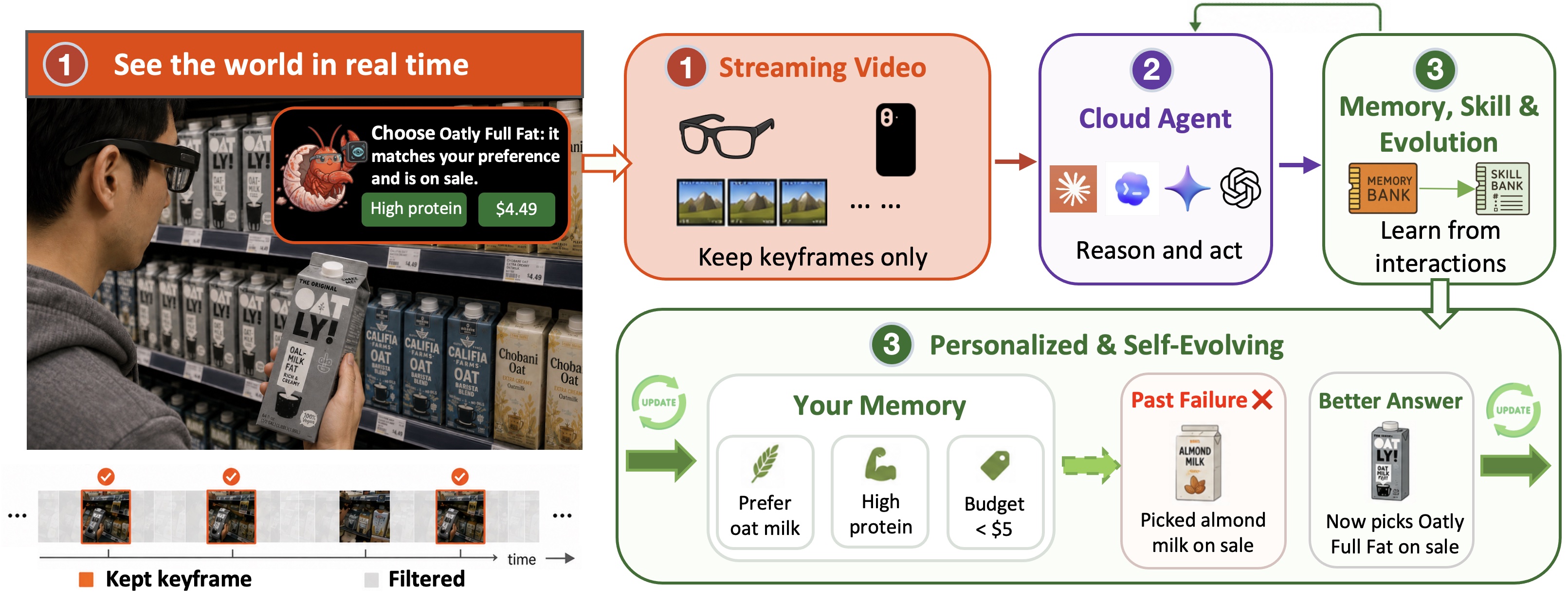}
\caption{\mmmc{} can efficiently encode streaming video in real scenarios and produce personalized answers and actions through constantly evolved memory and skill banks. It uses cascaded video encoding to compress streaming observations into critical visual evidence, then retrieves task-relevant skills and memories to ground answers and executable actions. 
By accumulating failure experiences, memory-guided evolution updates the agent over time, enabling efficient perception, continual evolution, and user-specific problem solving across multimodal agent tasks.}
\label{fig:teaser}
\end{figure*}

\begin{abstract}
Vision language models (VLMs) are serving as general-purpose interfaces for complex multimodal tasks. However, deployment still faces three gaps: VLMs typically incur high latency and cost when processing dense video frames and long prompts, the agent scaffold remains static after deployment, and standard video-QA benchmarks do not test whether agents can use visual evidence inside tool-using workspaces.
We present \textbf{\mmmc{}}, a self-evolving multimodal agent built around two principles. First, \emph{hybrid encoding} reduces deployment cost by filtering less informative streaming frames with a cascaded gate and compressing the text skill bank through hot/cold top-$k$ injection. 
Second, \emph{skill evolution} lets the agent learn from failures: retrieved memories condition an offline evolver either as direct concatenated context or as guided evidence, producing skill-bank updates that help future questions.
Across $4$ video-QA benchmarks with $2$ VLMs (Gemini 3 Flash and GPT-5.2), \mmmc{} cuts per-question API cost by an average $-98\%$ versus full-frame upload (peak $-99.3\%$ on Video-MME) and by $-25.9\%$ over the offline uniform 8 frame baseline under the same evolved skill bank, while boosting accuracy in most settings, \eg, an average $+3.85\%$ and a peak $+15.80\%$ on EgoSchema with Gemini 3 Flash.
To address the benchmark gap, we further curate \textbf{\mmbench{}}, a $200$-scenario multimodal agentic benchmark built through a strict five-stage pipeline; models must use video evidence, documents, dynamic updates, and executable checks inside a workspace.
On \mmbench{}, the same self-evolution framework with computer-use agent backends improves macro accuracy by $+2.9\%$ for Codex (GPT-5.5) and $+3.2\%$ for Claude Code (Sonnet 4.6) over no-evolution baselines, with a $-9.5\%$ cost reduction compared to the uniform-sampled baseline.
These properties make \mmmc{} a natural fit for live edge applications such as AI glasses, where the cascade reduces a $1$-hour streaming session from $\sim3,600$ API uploads down to only $5$-$20$ calls and the self-evolution makes it a perfect personalized assistant.
\end{abstract}

\section{Introduction}
\label{sec:intro}
Multimodal models are becoming the default interface for agents that combine perception, language, memory, and tool use. Vision-language models (VLMs) are a concrete subset: with full visual information and long prompts, frontier VLMs already perform well on standard video question-answering (video-QA) multiple-choice (MC) benchmarks~\citep{mangalam2023egoschema, fu2024videomme}.
However, this offline setup hides three deployment gaps: the full multimodal context is assumed to be available at query time, the model scaffold is assumed to stay fixed, and the evaluation is usually limited to static video-QA rather than workspace-style multimodal agents. Real video-centric agents, from cloud assistants to wearable AI glasses, break all three. Visual observations arrive as streams of unknown length; the agent should improve from its own failures; and practical tasks require inspecting files, reconciling video evidence with text records, editing workspace artifacts, and passing executable checks. VisionClaw~\citep{liu2026visionclaw} makes this setting concrete by combining Meta's AI glasses, live multimodal perception, and OpenClaw execution, but it leaves self-evolving skill/memory banks, prompt-cost growth, edge-side frame gating, and benchmarked agentic evaluation largely untested.

These gaps have so far been addressed mostly in isolation. 
On the \textit{efficiency} side, frame-selection systems~\citep{zhang2024llovi,song2024moviechat,he2024malmm,wang2024videoagent} compress or subsample the input video, but they usually assume offline access to the full clip and do not address prompt-side cost from a growing skill bank. 
On the \textit{adaptation} side, skill libraries for large language model (LLM) agents~\citep{wang2024voyager,zhao2024expel,tang2025agentkb,xia2026metaclaw} distill reusable behavioural rules from past failures and inject them at inference, demonstrating that frontier-model accuracy can be lifted \emph{without weight updates}. 
However, these systems remain text-only; in multimodal settings, a growing bank of skills and memories can become a new token bottleneck. On the \textit{evaluation} side, standard video-QA benchmarks test one-shot answering, so they barely measure whether an agent can use visual evidence while conducting real actions and satisfying automatic checks.

We present \mmmc{}, a self-evolving multimodal agent built around two principles. (i) \textit{Hybrid encoding} for efficient edge deployment. The video stream is filtered by a lightweight cascade: perceptual hash, 128-dim CPU encoder, and adaptive change gate, so only salient video frames are sent to the VLM API. 
The skill bank is also encoded efficiently: a small hot top-$k$ set is injected with full skill text, while the remaining skills are exposed as a compact cold catalogue. (ii) \textit{Self-evolving} skills and memories for adaptive agents. Correctly-solved examples are stored in the memory bank, and failures trigger an offline LLM evolver that uses relevant memories and failures to update the skill bank. Per-skill utility tracking and pruning keep the bank compact, enabling personalization and benchmark-specific adaptation without updating VLM weights. We show a running example of what \mmmc{} is capable of in Figure~\ref{fig:teaser}.

To address the third gap, we curate \mmbench{}, a $200$-scenario multimodal agentic benchmark for evaluating \mmmc{} under tool-using execution with multiple-choice questions and executable-check problems. 
Starting from various existing video data~\citep{yang2025thinking,mangalam2023egoschema,lei2021detecting}, we build documents, chat/audio traces, dynamic updates, and executable checkers, then filter scenarios with a five-stage curation pipeline: candidate generation, timestamp-grounded workspace construction, paired text-only/with-clip leakage filtering, multimodal criteria-based selection, and health checks. The final suite contains an average of $24.4$ steps and $18.1$ visual-required steps per scenario, making \mmbench{} a validated testbed for multimodal computer-use agents.

For static video-QA, we evaluate $4$ benchmarks and $2$ VLM families: two egocentric benchmarks (EgoSchema, EgoPlan-Bench) and two general-video benchmarks (Video-MME long, NextQA), tested on Gemini~3~Flash and GPT-5.2.
The streaming configuration with the proposed skill evolution improves accuracy in most settings, with an average lift of $+3.85\%$ and a peak of $+15.80\%$ on EgoSchema using Gemini~3~Flash. 
When the same skill evolution is added to a stronger offline Uniform-8 baseline, it contributes another $+3.50\%$ to $+13.00\%$.
On efficiency, hybrid encoding reduces per-question API cost by $-98.1\%$ against full-frame upload and by $-25.9\%$ against the Uniform-8 baseline with self-evolution; at a matched $K{=}8$ frame budget, cascade-fill still beats uniform sampling on short-clip benchmarks.
We further evaluate \mmmc{} on \mmbench{} with two tool-using agent backends, Codex (GPT-5.5) and Claude Code (Sonnet 4.6), over 200 \mmbench{} scenarios. 
\mmmc{} with simple memory-to-evolver concatenation reaches $54.3\%$ macro accuracy with Codex and $52.2\%$ with Claude Code, improving over matched no-evolution Cascade-8 baselines by $+2.9$ and $+3.2$ points and over Uniform-8 by $+4.0$ and $+8.2$ points. The lift is strongest on empirically hard scenarios ($+5.4$ Codex, $+5.3$ Claude Code), and the Claude Code Cascade-8 setting costs $-9.5\%$ less than Uniform-8.
Together, these results show that the frame gate and the evolving skill/memory scaffold improve both static video-QA and workspace-style multimodal agents, while remaining compatible with frozen VLM backbones.

\section{Related Work}
\label{sec:related}
\paragraph{Skill-based and memory-augmented agents.} A line of work augments agents with reusable skill libraries or external memory to improve performance without modifying model weights. Reflexion~\citep{shinn2023reflexion} stores verbal self-reflections in an episodic buffer; Voyager~\citep{wang2024voyager} incrementally builds a library of executable code skills from successful episodes; ExpeL~\citep{zhao2024expel} and Agent-KB~\citep{tang2025agentkb} distill cross-task experience into natural-language rules. Memory systems include MemGPT~\citep{packer2023memgpt}, Generative Agents~\citep{park2023generative}, Mem0~\citep{chhikara2025mem0}, and MemEvolve~\citep{zhang2025memevolve}. A shared limitation is that the skill library is treated as a static artefact, not coordinated with weight optimisation. MetaClaw~\citep{xia2026metaclaw} addresses this by coupling skill evolution with RL training;
There are also recent computer use agents that leverage memory systems and specialized components for GUI tasks~\citep{han2026vlaa,pointer2026gui}.
Our \mmmc{} freezes the model weights, relying instead on per-skill utility tracking and bank hygiene to keep the library quality-controlled across long evolution histories.

\paragraph{Continual and meta-learning.} Meta-learning frames learning as optimisation for fast adaptation to new tasks. RL$^2$~\citep{duan2016rl2}, PEARL~\citep{rakelly2019pearl}, and ProMP~\citep{rothfuss2019promp} demonstrate fast adaptation in robotic control with low-dimensional action spaces. Continual learning studies sequential task adaptation without forgetting~\citep{kirkpatrick2017ewc,chaudhry2019agem,zenke2017synaptic}. Online meta-learning relaxes the offline assumption. MetaClaw~\citep{xia2026metaclaw} extends this to LLM agents in a non-stationary task stream, with strict support/query separation and a versioning protocol; \mmmc{} inherits the protocol structure and applies it to video-QA failure distillation, with per-skill utility tracking serving the role of MetaClaw's stale-reward filtering.

\paragraph{Selected-frame and efficient video VLMs.} A growing line of work selects a small subset of frames before invoking the VLM, but the prevailing approach puts an LLM \emph{in the per-frame selection loop} or assumes offline access to the full clip. LM-planner / locator selectors~\citep{wang2024videoagent,yu2025framevoyager,yu2023sevila,wang2025videotree} score candidate frames against the question via an inner LLM call; memory-augmented compressors~\citep{song2024moviechat,he2024malmm} use sliding windows or learned memory queries to fold long video into a fixed budget; uniform / chunked stacks~\citep{zhang2024llovi,ren2024timechat,huang2024vtimellm,yang2023vid2seq,zhang2023videollama,zhang2025llavavideo} accept the full clip at a fixed frame budget, optionally prefixing time tokens for temporal grounding. VisionClaw~\citep{liu2026visionclaw} shares our always-on wearable motivation: it couples Meta Ray-Ban smart glasses with live multimodal perception and OpenClaw execution, showing that perception plus agentic action can reduce user interaction overhead. However, these systems are either offline selectors or fixed agentic stacks. Table~\ref{tab:framesel-compare} summarizes the comparison: \mmmc{}'s cascade is the only mechanism in this set that runs CPU-only on the edge, decides frame-by-frame as frames arrive, and uses no LLM in the selection loop. Orthogonally, \mmmc{} also evolves the reasoning bank and memory store across the task stream, while the prompt-side cost of prior systems is fixed at deployment time.

\begin{table}[t]
\centering
\caption{Selected-frame VLMs. Selector is the frame selection method. Online indicates if the method keep/skip decision frame-by-frame as frames arrive. Edge-CPU: selector runs on-device with no GPU. And existing multimodal agents are unable to evolve natively.}
\label{tab:framesel-compare}
\scriptsize
\setlength{\tabcolsep}{13pt}
\begin{tabular}{lcccc}
\toprule
Method & Selector & Online & Edge-CPU & Evolve \\
\midrule
LLoVi~\citep{zhang2024llovi}              & uniform chunks & \xmark  & \xmark  & \xmark \\
MovieChat~\citep{song2024moviechat}          & sliding window & \xmark  & \xmark  & \xmark \\
MA-LMM~\citep{he2024malmm}             & trained mem-Q  & \xmark  & \xmark  & \xmark \\
VideoAgent~\citep{wang2024videoagent}         & LLM planner     & \xmark  & \xmark  & \xmark \\
Frame-Voyager~\citep{yu2025framevoyager}      & LLM planner     & \xmark  & \xmark  & \xmark \\
SeViLA~\citep{yu2023sevila}             & BLIP-2 locator & \xmark  & \xmark  & \xmark \\
VideoTree~\citep{wang2025videotree}          & LLM tree-descent & \xmark & \xmark  & \xmark \\
TimeChat~\citep{ren2024timechat}           & uniform sampling       & \xmark  & \xmark  & \xmark \\
VTimeLLM~\citep{huang2024vtimellm}           & uniform sampling       & \xmark  & \xmark  & \xmark \\
Video-LLaMA~\citep{zhang2023videollama}        & uniform sampling       & \xmark  & \xmark  & \xmark \\
LLaVA-Video~\citep{zhang2025llavavideo}        & uniform sampling      & \xmark  & \xmark  & \xmark \\
VisionClaw~\citep{liu2026visionclaw}        & n/a      & \cmark  & \cmark  & \xmark \\
\midrule
\textbf{\mmmc{}}   & \textbf{cascade (heuristic)} & \textbf{\cmark} & \textbf{\cmark} & \textbf{\cmark} \\
\bottomrule
\end{tabular}
\end{table}

\paragraph{Egocentric and general video-QA benchmarks.} We evaluate our method on both egocentric and general multimodal data. EgoSchema~\citep{mangalam2023egoschema} is the standard 5-way MC benchmark for 3-min ego clips; EgoPlan-Bench~\citep{chen2023egoplan} targets ego planning from a single observation frame. Ego4D~\citep{grauman2022ego4d}, Charades-Ego~\citep{sigurdsson2018charadesego}, EgoThink~\citep{cheng2024egothink}, and VidEgoThink~\citep{cheng2024videgothink} provide complementary task formats. For general video, Video-MME~\citep{fu2024videomme} covers 12 task types across various durations, NextQA~\citep{xiao2021nextqa} tests causal and temporal reasoning on $\sim\!30$\,s YouTube clips, and IntentQA~\citep{li2023intentqa} targets intent reasoning. We report on EgoSchema, EgoPlan-Bench, Video-MME long, and NextQA.


\section{\mmmc{}: An Efficient Multimodal Agent that Evolves Itself}
\label{sec:method}

\begin{figure*}[h]
\centering
\includegraphics[width=\textwidth]{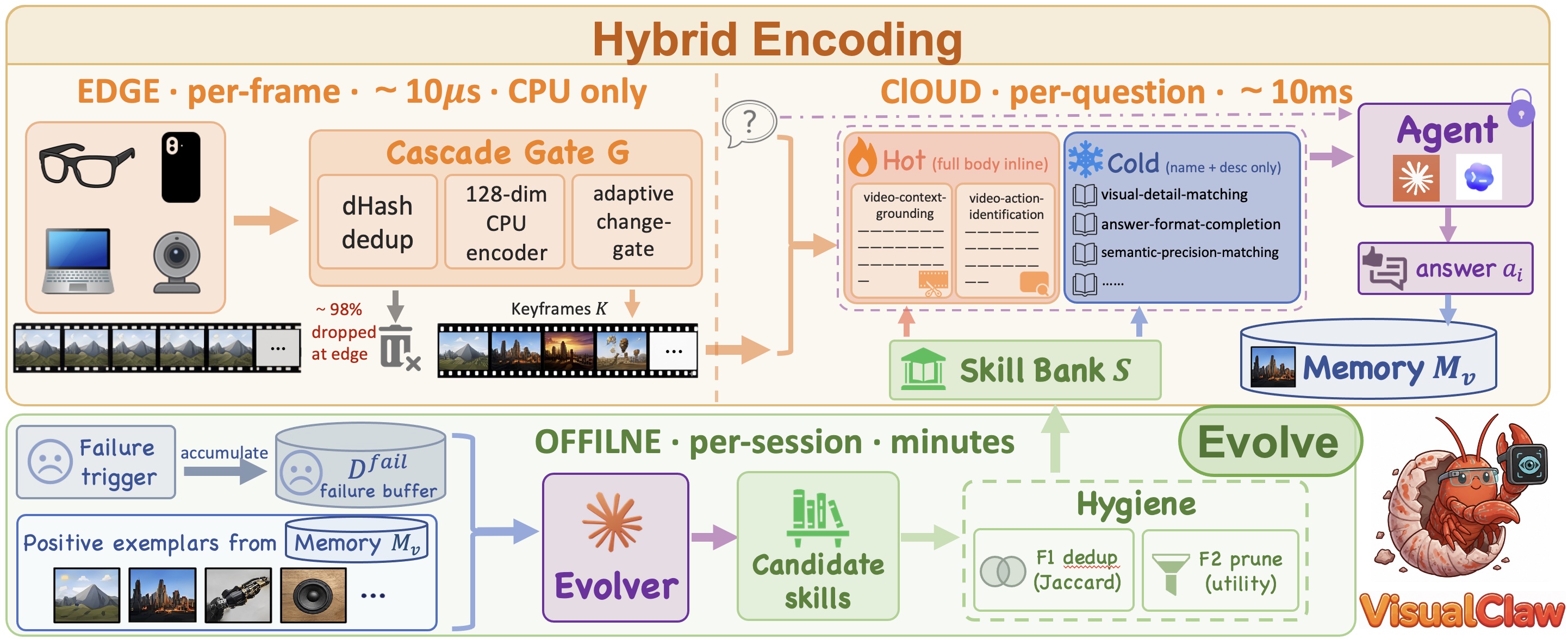}
\caption{
\textbf{\mmmc{} pipeline.} A three-timescale system with hybrid encoding and meta-evolve. An on-device cascade gate encodes frames \emph{per-frame}, and a memory-augmented evolver paired with a hot/cold skill injector evolves the language-layer scaffolding \emph{per-question} and \emph{per-session}.
}
\label{fig:pipeline}
\end{figure*}

The \mmmc{} framework addresses the first two deployment gaps: streaming video is too expensive to upload densely, and a fixed agent scaffold cannot adapt after deployment. It composes three filtering stages, each operating at its own timescale: an edge-side cascaded gate $G$ that triages frames \emph{per-frame}, a hot/cold skill injector that triages the bank $S$ \emph{per-question}, and a memory-augmented evolver that distils new entries into $S$ \emph{per-session} from a confidence-gated episodic store $M_v$. The first stage cuts the visual cost; the latter two evolve the language-level scaffolding $\{S, M_v\}$ over the task stream. VLM weights $\theta$ are never updated as we make VLM calls through API. 
We present the \mmmc{} pipeline of hybrid encoding and meta-evolve in Figure~\ref{fig:pipeline}.

\subsection{Preliminaries and Notation}
\label{sec:formalism}

We denote the VLM to evolve as $M = (\theta, S, M_v, G)$, with $\theta$ the (frozen) weights of a cloud VLM, $S=\{s_1,\dots,s_K\}$ a library of language skills, $M_v$ an episodic memory store indexed by dense sentence embeddings, and $G$ an edge-side cascade visual encoding gate we proposed.
For a question $q$ over a stream of frames $\mathcal{F}$, the answer $a$ is generated by:
\begin{equation}
\small
a \sim \pi_\theta\!\left(\cdot \mid q,\; G(\mathcal{F}),\; \mathrm{Ret}_S(q,k),\; \mathrm{Cat}(S),\; \mathrm{Ret}_{M_v}(q)\right),
\end{equation}
where $\mathrm{Ret}_S(q,k)$ returns the top-$k$ retrieved skills (the hot tier), $\mathrm{Cat}(S)$ produces the name-and-description cold catalogue, and $\mathrm{Ret}_{M_v}(q)$ retrieves confidence-gated memory snippets. The three components $\{G, S, M_v\}$ are updated at three qualitatively-different timescales:

\begin{tcolorbox}[colback=blue!5,colframe=blue!50,boxrule=0.4pt,arc=2pt]
\textbf{Per-frame ($\sim\!10\,\mu$s, edge).} $G$ inspects each arriving frame and emits a \textsc{major} / \textsc{minor} / \textsc{skip} verdict. CPU-only.\\[2pt] \textbf{Per-question ($\sim\!10$\,ms, cloud).} $\mathrm{Ret}_S$ and $\mathrm{Ret}_{M_v}$ rank the bank against $q$; the top-$k$ skills are inlined, the rest catalogued.\\[2pt] \textbf{Per-session (minutes, offline).} After every $N_{\text{evo}}$ failures the LLM evolver $\mathcal{E}$ analyses failure trajectories and proposes new skills, $S\leftarrow S\cup\mathcal{E}(S,D^{\text{fail}},M_v)$; bank-hygiene filters $F_1, F_2$ maintain bank quality.
\end{tcolorbox}
We also present the detailed algorithm in Appendix~\ref{app:detail-algo}.


\subsection{Per-frame: Cascaded Encoding Gate}
\label{sec:cascade}

Most frames within a contiguous window of a streaming video are visually redundant, and cloud-VLM cost grows linearly with each forwarded frame.
We therefore filter the stream content-aware on the edge, forwarding only the salient transitions with a decision rule that consumes no future frames --- ruling out any selector that needs the full clip up front.
Concretely, the visual encoding gate $G$ maps each arriving frame $f_t$ to a verdict $g_t\in\{\textsc{major}, \textsc{minor}, \textsc{skip}\}$ from $f_{1:t}$ alone: \textsc{major} frames cross the major-change threshold and are forwarded to the cloud VLM as keyframes, \textsc{minor} frames cross only a lower threshold and update the rolling reference for subsequent comparisons but are not uploaded, and \textsc{skip} frames fall below both thresholds and trigger no action. The keyframe set forwarded to the cloud is therefore $K=\{f_t : g_t=\textsc{major}\}$. $G$ composes three $O(1)$-per-frame stages applied in order: a perceptual hash (dHash) that Hamming-dedups against a rolling buffer to drop bit-exact and near-exact duplicates (camera shake, stationary periods); a lightweight 128-dim CPU encoder (HSV histogram, luminance, edge density, texture) that produces a cosine-comparable scene vector with no deep network or GPU; and an adaptive change gate that compares the encoded frame against a rolling reference and emits \textsc{major}/\textsc{minor}/\textsc{skip} verdicts under temporally-decaying thresholds, so the gate fires reliably on slow-moving scenes and stationary cameras. Because $g_t$ depends only on $f_{1:t}$, the cascade runs on a live stream of unknown length without buffering or replay; only frames in $K$ are forwarded to the cloud, everything else is discarded at the edge.

\subsection{Per-question: Skill Bank with Hybrid Hot/Cold Injection}
\label{sec:skillbank}
Agent scaffolds empower LLM agents without weight updates, but each injected skill costs prompt tokens at every query, and once the skill bank $S$ exceeds tens of entries full-inline injection saturates the prompt context and obscures task-specific signal.
We therefore inject the skill in two tiers.
Each text skill $s_j\in S$ is a short markdown card with a name, a one-line description (the retrieval key for skill evolve), a numbered procedural body, and an explicit anti-pattern section, and $S$ is initialized with a small seed bank of $K_{\text{seed}}$ cross-cutting visual-reasoning patterns that are bootstrapped from a held-out probe run of the same skill evolver $\mathcal{E}$, and grows during deployment thereafter.
We treat each $s_j$ as an \emph{implicit preference rule} rather than a procedural recipe.

To adopt skills dynamically, for each incoming question $q$, a sentence-transformer embedding ranks $S$ against the question text. The top-$k$ skills $S^{\text{hot}}=\mathrm{Ret}_S(q,k)$ are inlined as \emph{hot} bodies into the system prompt; the remaining skills become a \emph{cold catalogue} $S^{\text{cold}}=\mathrm{Cat}(S\setminus S^{\text{hot}})$ of name-and-description pairs only, with bodies fetchable on demand if the model decides it needs them. Per-question prompt cost is therefore bounded by $k$ rather than $|S|$, decoupling injection cost from bank growth.

\subsection{Per-session: Memory-Augmented Meta-Evolution}
\label{sec:memevolve}
As use cases become complex in real-life, a static skill bank $S$ cannot cover unseen failure modes. We therefore use an LLM evolver $\mathcal{E}$ that updates $S$ from recent failures while using memory to avoid narrow, failure-specific recipes and near-duplicate skill growth.

Prior memory-augmented agents often concatenate retrieved examples into the current model prompt~\citep{shinn2023reflexion,packer2023memgpt,xia2026metaclaw}. We keep this idea, but apply it to the lower-frequency skill-evolution prompt. The memory store $M_v$ saves correctly-answered examples as dense embeddings and retrieves only high-confidence matches. Let $D^{\text{fail}}$ be the failures since the last evolution. When $|D^{\text{fail}}|\geq N_{\text{evo}}$, the evolver retrieves relevant memory from $M_v$ and proposes new skills: $\Delta S=\mathcal{E}(S,D^{\text{fail}},M_v)$.

We instantiate the memory-to-evolver channel in two ways, reported as \texttt{Cat.} and \texttt{Guide} in Tables~\ref{tab:main} and~\ref{tab:mm-arena}. Let $R=\mathrm{Ret}_{M_v}(D^{\text{fail}})$ denote the bounded memory retrieved against the current failure batch, and let $\mathcal{P}_0(S,D^{\text{fail}})$ be the standard skill-evolution prompt. The \texttt{Cat.} variant follows the same direct concatenation spirit:
\begin{equation}
\small
\Delta S_{\texttt{Cat.}}=\mathcal{E}\!\left(\mathcal{P}_0(S,D^{\text{fail}})\oplus R\right),
\end{equation}
where the retrieved memory is appended as auxiliary context without changing the evolver instruction. The \texttt{Guide} variant adds a lightweight prefix that tells the evolver how to use the retrieved memory:
\begin{equation}
\small
\Delta S_{\texttt{Guide}}=\mathcal{E}\!\left(\mathcal{P}_{\text{guide}}(R)\oplus \mathcal{P}_0(S,D^{\text{fail}})\right),
\end{equation}
where $\mathcal{P}_{\text{guide}}$ marks the retrieved examples as failure-related context and instructs the evolver to extract reusable skills while avoiding scenario-specific details. Both variants differ from \texttt{+SkillMemCat}: memory is still concatenated or guided, but it affects the next skill update rather than the high-frequency per-question VLM prompt.

Two filters keep the skill bank $S$ quality-controlled across long evolution histories: (F1) a token-Jaccard dedup at evolve-time rejects entries in $\Delta S$ whose names overlap heavily with existing $s_j\in S$, and (F2) a per-skill utility tracker logs each $s_j$'s hit rate on scored answers and periodically prunes skills whose accuracy lags the bank mean. 
Together they make the per-session evolution loop ``self-healing'' rather than monotonically blowing.


\section{\mmbench{}: A Multimodal Agentic Benchmark}
\label{sec:mmbench}
Static video-QA benchmarks test one-shot answer selection, leaving the third deployment gap: they do not measure whether a multimodal system can use visual evidence while conducting real computer operations, \eg, editing files, maintaining workspace state, and recovering from previous mistakes. 
We therefore introduce \mmbench{}, an agentic benchmark that wraps tool-using backends such as Codex and Claude Code behind the same staged-workspace interface.

\begin{figure*}[h]
\centering
\includegraphics[width=\textwidth]{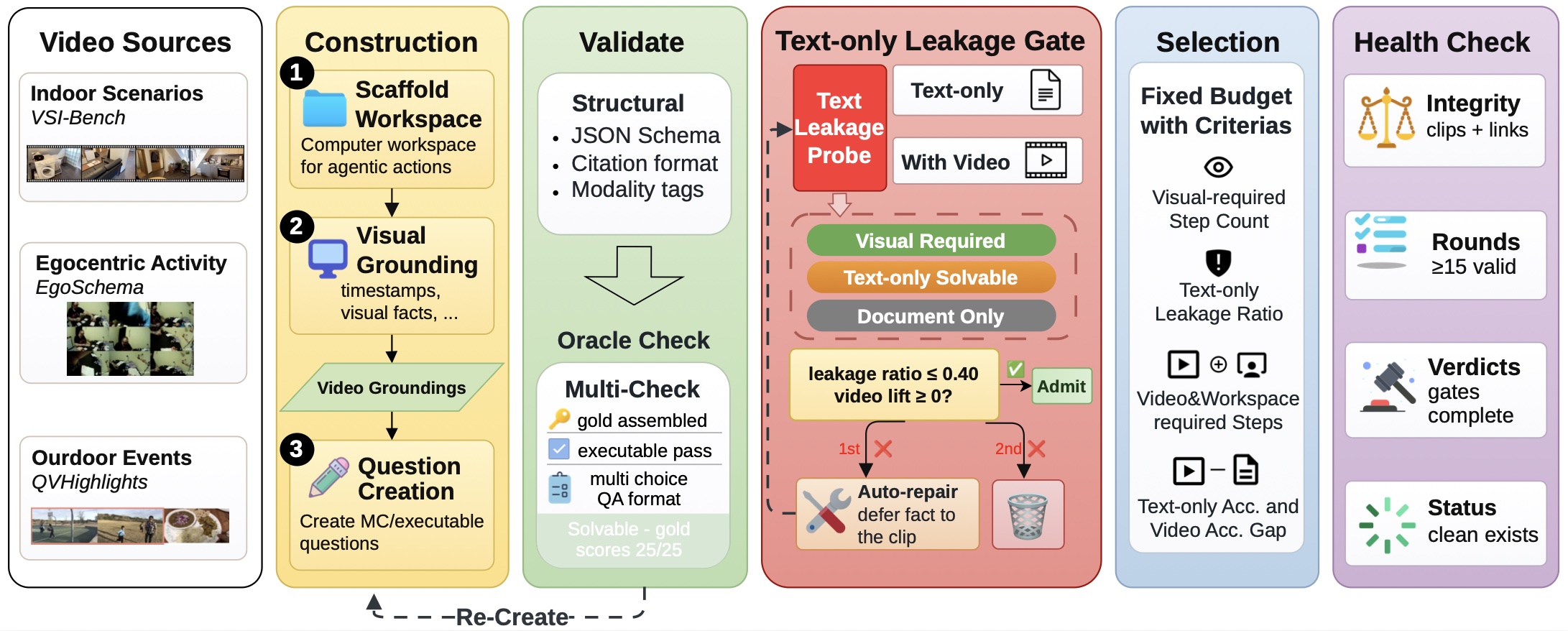}
\caption{
\textbf{\mmbench{} curation pipeline.} We choose videos from three different sources and apply five strict data processing stages, including construction, filtering process to make sure the curated data are executable, of high-quality, and diverse.
}
\label{fig:mmbench-curation}
\end{figure*}

\subsection{Building \mmbench{}}
We build our \mmbench{} upon $200$ multimodal scenarios with video from existing datasets: $100$ Indoor/VSI~\citep{yang2025thinking}, $50$ EgoSchema~\citep{mangalam2023egoschema}, and $50$ QVHighlights~\citep{lei2021detecting} videos. 
Each scenario pairs a short video with a workspace of documents, chat/audio traces, dynamic updates, and executable checks. Agents must read and edit files, reconcile contradictions between the video and text records, and leave a final workspace that can be automatically scored rather than emit a bare answer. We present the data curation pipeline in Figure~\ref{fig:mmbench-curation}.

\begin{figure*}[t]
\centering
\includegraphics[width=.9\textwidth]{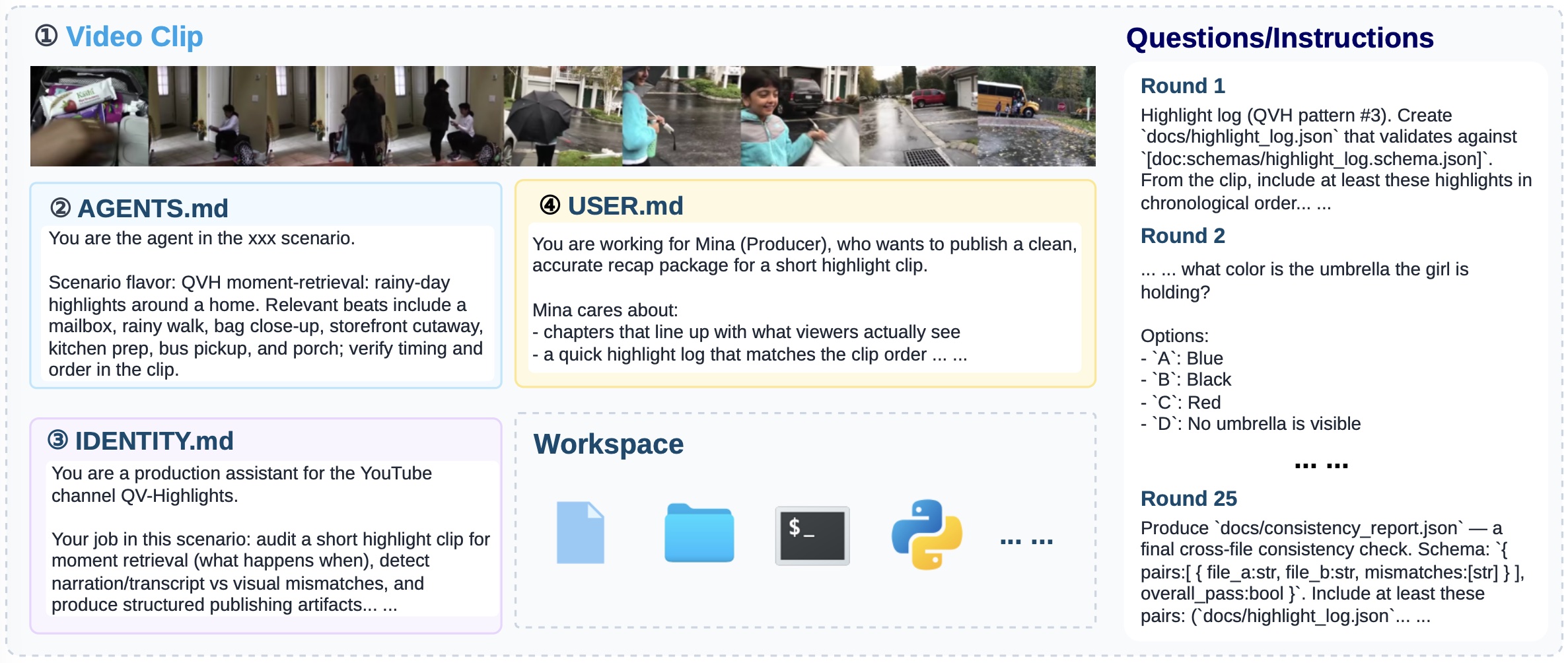}
\caption{A complete example from \mmbench{} with an video clip, scenario-related files, and the workspace. The questions/instructions of each evolving step in \mmbench{} include both multi-choice questions and computer-level operational requests.}
\label{fig:mmbench-case}
\end{figure*}

\paragraph{Input and Construction.}
For each candidate, we set up a computer-use workspace with the clip, background facts used for grading, documents, chat/audio records, updates, and metrics scripts.
By default, each scenario follows the same file scaffold as follows:
\begin{itemize}[leftmargin=1.2em,itemsep=1pt,topsep=2pt]
\item \texttt{clip.mp4}: the source video with selected visual evidence frames available to the agent.
\item \texttt{AGENTS.md}: the operating contract, including the task goal, available evidence, output protocol, citation rules, and scenario-level constraints.
\item \texttt{IDENTITY.md}: a compact role card that specifies who the agent is and what background assumptions it should keep fixed.
\item \texttt{USER.md}: the stakeholder request and intent, so the task is grounded as a real workspace need rather than a bare QA prompt.
\item \texttt{questions.json} and \texttt{check\_*.py}: the round instructions and executable checkers used to score the final workspace state.
\end{itemize}
We then ask the Gemini 3.1 Pro to mark which timestamps support important visual facts, so later questions cannot cite visual evidence that is not actually visible. 
The original datasets provide videos, not our agentic questions. Given the timestamped visual facts and workspace state, systems create new scenario steps, workspace files, updates, reference answers, and scoring scripts.
The resulting tasks include multiple-choice visual checks and executable workspace tasks, such as writing evidence tables, updating JSON/CSV files, reconciling stale documents with the clip, and applying dynamic updates across later steps. We show an example from QVHighlights in Figure~\ref{fig:mmbench-case}.

\paragraph{Validation.}
We first check if each scenario is well-formed from several aspects and make sure that: 1. the JSON schema is valid (required fields, multiple-choice options, and answer keys are structurally consistent); 2. citations follow the required timestamp format, every step declares the evidence modalities it needs (\eg, video, image, text, audio); 3. and at least 40\% of steps require video, so the suite is visually grounded by construction. 
We then assemble the gold solution into a fresh workspace to examine if the environment works correctly (\eg, every executable check passes and every multiple-choice step is well-formed). 
Finally, scenarios that are malformed or whose reference solution does not pass are rewritten at the end.
\paragraph{Text-only Leakage Gate.}
We test each candidate twice with the same VLM (\eg, Gemini 3.1 Pro): once with the clip hidden, and once with the clip frames available. 
If a step can be solved without the clip, it is not a true visual step. We label steps as \texttt{visual\_required}, \texttt{text\_only\_solvable}, or \texttt{doc\_only}. A data candidate is kept when fewer than 40\% of its steps are solvable from text alone, and the with-clip run is not worse than the text-only run (\ie, $\Delta \geq 0$). Otherwise, we repair it once by moving the decisive clue back into the video evidence.

\paragraph{Selection.}
To build a high-quality benchmark from the curated examples, we rank surviving candidates with a heuristic weighted score, $s=v+8\max(\Delta,0)-6L+0.5m$. 
Here $v$ is the number of visual-required steps, $L$ is the text-only leakage ratio, and $m$ counts steps that require both video and another workspace source. 
We obtain $\Delta$ from the paired leakage probe above: the same fixed VLM answers the same steps once with the clip hidden and once with the clip available, and $\Delta=\text{Acc}_{\text{clip}}-\text{Acc}_{\text{text}}$. 
The weights make visual-required coverage the main signal, reward cases where the clip truly helps, penalize text leakage, and use cross-modal steps as a small bonus. 

\paragraph{Health Check.}
The final pass checks that all video clips are present, each scenario has at least $15$ valid steps, and every step has a leakage label. 
The Table~\ref{tab:mmbench-stats} summarizes the final suite; we present all the main results on \mmbench{} with $200$ scenarios and $3{,}106$ steps in total so experiment statistics are directly comparable.

\begin{table}[t]
\centering
\caption{\mmbench{} statistics. MC/Exec gives the average number of multiple-choice and executable-check steps per scenario. VR Steps means video-required steps per scenario. Easy/Med./Hard are based on the average Cascade-8 w/o \texttt{FullEvo} accuracy over Codex and Claude Code backend \mmmc{}.}
\label{tab:mmbench-stats}
\scriptsize
\setlength{\tabcolsep}{4pt}
\resizebox{.95\columnwidth}{!}{%
\begin{tabular}{lrrrrrr}
\toprule
Source & Scen. & Video Len. & Avg. Steps & Avg. MC/Exec & Avg. VR Steps  & Easy/Med./Hard \\
\midrule
VSI~\citep{yang2025thinking} & $100$ & $24$--$434$\,s & $23.9$ & $6.7/17.2$ & $15.4$ & $2/15/83$ \\
EgoSchema~\citep{mangalam2023egoschema} & $50$ & $180$\,s & $25.0$ & $7.3/17.7$ & $21.2$ & $39/10/1$ \\
QVHighlights~\citep{lei2021detecting} & $50$ & $121$--$150$\,s & $24.9$ & $7.1/17.9$ & $20.4$ & $4/21/25$ \\
\midrule
Total & $200$ & $24$--$434$\,s & $24.4$ & $6.9/17.5$ & $18.1$ & $45/46/109$ \\
\bottomrule
\end{tabular}
}
\end{table}

\subsection{Bridge to Multimodal Agentic Benchmarks}
\label{sec:agent-bridge}
For each evolve step, the runner initializes a clean computer workspace, applies the step's document or data updates, stages the selected visual keyframes as local image files, and prepends retrieved skills or memory context to the agent prompt. The agent then acts in the workspace: it can inspect files and keyframes, edit outputs, and run lightweight checks before stopping.

Both backends share the same scoring contract but differ in execution. Claude Code runs as a native tool-using workspace agent, so reads, edits, and validation commands happen through its regular tool loop. Codex uses a stricter noninteractive bridge: the runner provides the workspace snapshot and image attachments up front, then applies the file writes emitted by Codex after the call returns. In both cases, scoring remains outside the agent. The evaluator checks the final workspace state for execution-style tasks or parses the final answer for multiple-choice rounds, and failures are routed back into the same memory and skill-evolution loop. This bridge tests whether the evolved skill bank helps not only video answer selection, but also multimodal agentic tasks that require visual grounding, file manipulation, and persistent workspace state.

\section{Experiments}
\label{sec:results}

\subsection{Setup}
\label{sec:setup}

\textbf{Evaluation Setup.}
We evaluate our \mmmc{} framework on two types of tasks: traditional video-QA and the proposed multimodal agentic benchmark \mmbench{}.
For static video-QA, we test Gemini~3~Flash~\citep{google2025gemini3flash} and GPT-5.2~\citep{openai2025gpt52} as frozen VLM backbones.
For \mmbench{}, we evaluate two tool-using backends, Codex and Claude Code, using the staged-workspace bridge in Sec.~\ref{sec:agent-bridge}.
We use Claude Haiku 4.5~\citep{anthropic2025haiku45} as the offline evolver for video-QA and \texttt{all-MiniLM-L6-v2}~\citep{reimers2019sentencebert} as the memory encoder.
To adapt to streaming video setting, for both task types, videos are sampled at $1$\,fps and each query is capped at $8$ keyframes. The cascade uses a major-change gate with $\tau_{\text{major}}{=}0.30$ and a $10$\,s silence ceiling. Unless otherwise specified, $N_{\text{evo}}{=}15$ failures trigger evolution and $K_{\text{seed}}{=}12$ skills initialize the bank.
As for baseline methods, for video-QA, we compare \texttt{Plain}, \texttt{Seed}, \texttt{+Evolve}, \texttt{+SkillMemCat}, and the two \texttt{FullEvo} variants shown as (\texttt{Cat.}) and (\texttt{Guide}) in Table~\ref{tab:main}. For \mmbench{}, we report \mmmc{} with \texttt{Cat.}/\texttt{Guide}, \mmmc{} w/o \texttt{FullEvo}, and Uniform-8 baselines with or without the \texttt{FullEvo}.

\paragraph{Benchmarks.} For static video-QA, we evaluate \mmmc{} on two ego-view and two general video-QA datasets to simulate the use of real-life scenarios and to probe in more general multimodal situations.
For the ego-view datasets, we use EgoSchema~\citep{mangalam2023egoschema} with 500 instances, 3-min clips on average to test long-horizon ego activity recognition, and EgoPlan-Bench~\citep{chen2023egoplan} with 923 entries and single observation frame for planning tasks.
As for the general video-QA, we test on Video-MME long~\citep{fu2024videomme} with 900, and $30$+minute video content to test the reasoning at a long clip duration, as well as NextQA~\citep{xiao2021nextqa} with 1000 instances and $\sim$30s videos for short-clip causal/temporal reasoning.
For agentic evaluation, the proposed \mmbench{} contains $200$ visual scenarios with a total of $3{,}106$-round of instructions; macro scenario accuracy is the primary metric.


\subsection{Accuracy Results and Analysis on video-QA Tasks}
\label{sec:headline}

\paragraph{Static video-QA Gains Across Backbones.} In Table~\ref{tab:main}, the (\texttt{Guide}) column, \ie, \texttt{FullEvo} with guided memory-to-evolver context, achieves an average performance boost of $+3.85\%$ over the Plain baseline and a peak of $+15.80\%$ on EgoSchema with Gemini~3~Flash.
Notably, Gemini 3 Flash yields at least $4\%$ lift at the streaming budget on two egocentric benchmarks (EgoSchema $+15.80\%$, EgoPlan-Bench $+4.23\%$), demonstrating its advantage in real-world applications.

To better understand the role of each function, we ablate each component in Table~\ref{tab:main}. The initial skill bank (\texttt{Seed}) already lifts Gemini's Plain by $+6.10\%$ on average, indicating that the bootstrapped seed addresses the dominant Gemini failure modes. Adding a simple evolver without memory (\texttt{+Evolve}, $+5.73\%$), answer-time memory concatenation (\texttt{+SkillMemCat}, $+5.39\%$), or raw memory-to-evolver concatenation, the (\texttt{Cat.}) column ($+5.67\%$), remains close to but below \texttt{Seed} on Gemini as full context might cause burden for a single VLM call, and none of these three variants boosts GPT-5.2 above the Plain on average. The guided variant, the (\texttt{Guide}) column, is the most stable full-evolution setting: it beats \texttt{Seed} by $+0.33\%$ on Gemini and $+1.31\%$ on GPT-5.2, and is the only cascade variant that lifts the stronger GPT-5.2 backbone above Plain on average ($+1.27\%$ vs.\ $-0.04\%$ to $-0.62\%$ for the other four non-Plain variants).

\begin{table}[t]
\centering
\caption{Results of Gemini 3 Flash and GPT-5.2 across 4 benchmarks. The Cascade columns report 6 streaming configurations; Uniform-8 plain is the offline baseline upper-bound. The (\texttt{Cat.}) and (\texttt{Guide}) columns are the two \texttt{FullEvo} memory-to-evolver variants defined in Sec.~\ref{sec:memevolve}. We mark best per row across the cascade columns in \textbf{bold}.}
\label{tab:main}
\small
\setlength{\tabcolsep}{4pt}
\begin{tabular}{llcccc>{\columncolor{gray!10}}c>{\columncolor{gray!10}}c|c}
\toprule
& & \multicolumn{6}{c}{Cascade Encoding} & {Uniform-8} \\
\cmidrule(lr){3-8}\cmidrule(lr){9-9}
Benchmark & Model & Plain & Seed & \texttt{+Evolve} & \texttt{+SkillMemCat} & (\texttt{Cat.}) & (\texttt{Guide}) & Plain \\
\midrule
\multirow{2}{*}{EgoSchema}
  & Gemini  & 52.60 & 67.20 & 68.00 & 65.20 & 64.60 & \textbf{68.40} & 60.60 \\
  & GPT-5.2 & 64.00 & 66.60 & 66.20 & 67.20 & 65.60 & \textbf{68.00}   & 70.60 \\
\midrule
\multirow{2}{*}{V-MME long}
  & Gemini  & 60.33 & 61.56 & 61.33 & 62.78 & 62.56 & \textbf{64.22}  & 61.44 \\
  & GPT-5.2 & 55.89 & 54.00 & 52.22 & 54.67 & 52.78 & \textbf{55.89} & 58.78 \\
\midrule
\multirow{2}{*}{EgoPlan-Bench}
  & Gemini  & 24.62 & \textbf{30.80} & 29.93 & 28.31 & 30.04 & 28.85        & 37.96 \\
  & GPT-5.2 & 28.42 & 28.74 & 28.31 & 28.09 & 28.85 & \textbf{29.39}       & 43.06 \\
\midrule
\multirow{2}{*}{NextQA}
  & Gemini  & 72.70 & 75.10 & 73.90 & 75.50 & \textbf{75.70} & 74.50        & 77.70 \\
  & GPT-5.2 & 73.20 & 72.00 & 72.30 & 70.90 & 72.50 & \textbf{73.30}        & 78.90 \\
\bottomrule
\end{tabular}
\end{table}

\begin{figure}[t]
\begin{minipage}[t]{0.48\linewidth}
\makeatletter\def\@captype{table}
\setlength\tabcolsep{3pt}
\caption{Uniform-8 + \texttt{FullEvo} (\texttt{Guide}) delivers a positive lift across all four benchmarks using Gemini~3~Flash.}
\label{tab:offline-ceiling-gemini}
\scriptsize
\resizebox{\textwidth}{!}{\begin{tabular}{lccc}
\toprule
Bench & U-8 plain & U-8 + \texttt{FullEvo} (\texttt{Guide}) & $\Delta$ \\
\midrule
EgoSchema       & 60.60 & \textbf{73.60} & $\textcolor{blue}{+13.00}$ \\
V-MME long      & 61.44 & \textbf{66.78} & $\textcolor{blue}{+5.34}$ \\
EgoPlan-Bench   & 37.96 & \textbf{50.11} & $\textcolor{blue}{+12.15}$ \\
NextQA          & 77.70 & \textbf{81.20} & $\textcolor{blue}{+3.50}$ \\
\bottomrule
\end{tabular}
}
\end{minipage}
\hspace{.8pt}
\begin{minipage}[t]{0.48\linewidth}
\centering
\makeatletter\def\@captype{table}
\setlength\tabcolsep{2.5pt}
\caption{EgoSchema leaderboard results. All baselines are reproduced from cited works.}
\label{tab:sota}
\scriptsize
\resizebox{.9\textwidth}{!}{\begin{tabular}{lcc}
\toprule
Method & Acc.\,(\%) & Backbone \\
\midrule
Frozen Bilinear~\citep{mangalam2023egoschema} & 17.6 & ResNet+BERT \\
LongViViT~\citep{papalampidi2024longvivit} & 56.8 & ViViT \\
LLoVi (GPT-4o)~\citep{zhang2024llovi} & 67.6 & GPT-4o \\
VideoAgent~\citep{wang2024videoagent} & 60.2 & GPT-4 + LLM planner \\
Gemini 1.5 Pro~\citep{geminiteam2024gemini15} & 72.2 & Gemini 1.5 Pro \\
\midrule
\mmmc{} (streaming \texttt{FullEvo} (\texttt{Guide})) & 68.4 & Gemini 3 Flash \\
\mmmc{} (U-8+\texttt{FullEvo} (\texttt{Guide})) & \textbf{73.6} & Gemini 3 Flash \\
\bottomrule
\end{tabular}
}
\end{minipage}
\end{figure}

\paragraph{Validating \texttt{FullEvo} on the Offline Baseline.}
As the increased accuracy of \mmmc{} primarily comes from the memory-driven skill evolution we proposed, we apply \texttt{FullEvo} to the Uniform-8 offline method in Table~\ref{tab:offline-ceiling-gemini} to better validate its use.
Even added to the stronger baseline, Uniform-8 + \texttt{FullEvo} (\texttt{Guide}) delivers an additional $+3.50$ to $+13.00\%$ over the offline Plain baseline, peaking at $+13.0\%$ on EgoSchema ($60.6\%\!\to\!73.6\%$) and $+12.15\%$ on EgoPlan-Bench.
This forms an offline upperbound that our streaming (\texttt{Guide}) variant approximates within $5.20\%$ on EgoSchema and $2.56\%$ on V-MME long while running at a small fraction of the cost (see Sec.~\ref{sec:efficiency}).

\begin{wraptable}{r}{0.52\columnwidth}
\vspace{-1.0em}
\centering
\caption{Parity-budget comparison at matched frame budget ($K{=}8$, Gemini~3~Flash).}
\label{tab:cascade-fill-headline}
\small
\setlength{\tabcolsep}{3pt}
\renewcommand{\arraystretch}{1.05}
\begin{tabular}{lcc}
\toprule
Method & NextQA & EgoPlan-Bench \\
\midrule
U-8
& 77.70
& 37.96 \\
Cascade-fill
& 79.50 {\tiny$_{\textcolor{blue}{+1.80}}$}
& 41.54 {\tiny$_{\textcolor{blue}{+3.58}}$} \\
\midrule
U-8 + \texttt{FullEvo}
& 81.20
& 50.11 \\
Cascade-fill + \texttt{FullEvo}
& \textbf{82.70} {\tiny$_{\textcolor{blue}{+1.50}}$}
& \textbf{52.06} {\tiny$_{\textcolor{blue}{+1.95}}$} \\
\bottomrule
\end{tabular}
\end{wraptable}

\paragraph{Cascade-fill Beats Uniform-8 at Matched Frame Budget.}
Table~\ref{tab:cascade-fill-headline} isolates frame-selection quality from frame count. On short-clip benchmarks (\eg, EgoPlan-Bench and NextQA, both with $\leq\!30$\,s clips), our cascade naturally selects only $1.13$-$1.51$ keyframes per question, below the Uniform-8 budget. We therefore test \emph{cascade-fill}: keep all cascade-selected keyframes and pad up to $K{=}8$ with uniformly-sampled fillers.
At this matched $K{=}8$ budget, cascade-fill beats pure uniform-$8$ both without skill scaffolding ($+1.80\%$ on NextQA, $+3.58\%$ on EgoPlan-Bench) and with \texttt{FullEvo} (\texttt{Guide}) on top ($+1.50\%$ on NextQA, $+1.95\%$ on EgoPlan-Bench), showing that scene-change keyframes carry more signal than evenly-spaced ones at fixed cost. Cascade-fill + \texttt{FullEvo} (\texttt{Guide}) further reaches \textbf{$52.06\%$} on EgoPlan-Bench, a $+14.10\%$ lift over plain U-$8$ ($37.96$), so frame quality and skill evolution compound rather than compete.

\paragraph{Guided Memory Evolution Fits Static video-QA.}
We further compare the two memory-to-evolver variants. \texttt{+SkillMemCat} concatenates retrieved memory into the high-frequency answer prompt and disables the evolver. In contrast, both (\texttt{Cat.}) and (\texttt{Guide}) send memory to the lower-frequency evolver; (\texttt{Cat.}) appends raw retrieved examples, while (\texttt{Guide}) adds an instruction prefix that asks the evolver to abstract reusable skills and avoid scenario-specific details.
In Table~\ref{tab:main}, (\texttt{Guide}) beats (\texttt{Cat.}) in six of eight rows, with an average margin of $+1.24\%$ and a peak of $+3.80\%$ on EgoSchema with Gemini~3~Flash. This is expected for static video-QA: each question is a single MC decision, so raw retrieved examples can become narrow context that the model has only one pass to use. The guided prefix pushes the evolver to convert those examples into compact visual-reasoning skills, which transfers better across VLM families, especially on GPT-5.2 where (\texttt{Guide}) wins every benchmark.
However, when it comes to agent systems that can use long prompts more smartly, this preference may change: workspace-style backends can process longer context through tool calls, edits, and verification loops. In such settings, raw memories may serve as concrete procedural traces, while static video-QA mainly rewards compact abstracted skills.

\paragraph{\mmmc{} Surpasses Frontier Baselines on EgoSchema.} On the EgoSchema leaderboard (Table~\ref{tab:sota}), \mmmc{}'s streaming \texttt{FullEvo} (\texttt{Guide}) setting reaches \textbf{$68.40\%$}, beating VideoAgent's offline LLM-driven planner ($60.20\%$) by $8.20\%$ at a fraction of the latency since the cascade rejects $\sim\!98\%$ of frames before any radio is woken (Sec.~\ref{sec:efficiency}); pushing to the offline configuration (Uniform-8+\texttt{FullEvo} (\texttt{Guide}), \textbf{$73.60\%$}) edges past Gemini~1.5~Pro's $72.20\%$ on a $4\times$ smaller, cheaper Gemini~3~Flash backbone and outperforms LLoVi (GPT-4o) by $6.00\%$ ($73.60\%$ vs.\ $67.60\%$), showing the gain compounds when the offline budget is available.

We show more quantitative diagnostics in Appendix~\ref{app:perf-analysis}, including capability-conditioned transfer, skill-bank and memory dynamics, and the hot/cold top-$k$ injection trade-off, which support the same design choices of our \mmmc{}.


\subsection{Results and Analysis on \mmbench{}}
\label{sec:mm-arena}

Using the \mmbench{} suite defined in Sec.~\ref{sec:mmbench}, we evaluate whether the same self-evolution stack transfers from video-QA to multimodal agentic work. Macro accuracy is primary because it gives each scenario equal weight; Table~\ref{tab:mm-arena} reports early/mid/late macro accuracy over within-scenario steps.


\begin{table*}[t]
\centering
\caption{\mmbench{} results. We compare two memory-to-evolver context modes for \mmmc{}: retrieved failures concatenated with skills (\texttt{Cat.}) and memory-guided skill evolution (\texttt{Guide}), plus \mmmc{} w/o \texttt{FullEvo} and Uniform-8 baselines with or without \texttt{FullEvo}. In long-horizon complex multimodal agent tasks, \mmmc{} (\texttt{Cat.}) achieves the best performance.}
\label{tab:mm-arena}
\small
\setlength{\tabcolsep}{8pt}
\begin{tabular}{llrrr>{\columncolor{gray!10}}r>{\columncolor{gray!10}}r}
\toprule
Backend & Setting & Early & Mid & Late & Micro & Macro \\
\midrule
\multirow{5}{*}{Codex}
 & \mmmc{} (\texttt{Cat.}) & 49.69 & \textbf{56.19} & \textbf{57.38} & \textbf{59.88} & \textbf{54.27} \\
 & \mmmc{} (\texttt{Guide}) & \textbf{50.66} & 54.66 & 56.88 & 59.50 & 53.89 \\
 & \mmmc{} w/o \texttt{FullEvo} & 48.10 & 52.93 & 53.47 & 57.95 & 51.35 \\
 \cmidrule{2-7}
 & Uniform-8 & 46.74 & 50.49 & 53.90 & 56.08 & 50.25 \\
 & Uniform-8 w/o \texttt{FullEvo} & 44.35 & 49.04 & 52.58 & 55.15 & 48.51 \\
\midrule
\multirow{5}{*}{Claude Code}
 & \mmmc{} (\texttt{Cat.}) & \textbf{52.03} & \textbf{50.91} & \textbf{53.50} & \textbf{57.79} & \textbf{52.16} \\
 & \mmmc{} (\texttt{Guide}) & 50.87 & 49.98 & 51.40 & 56.54 & 50.77 \\
 & \mmmc{} w/o \texttt{FullEvo} & 48.80 & 48.51 & 49.68 & 55.34 & 49.00 \\
\cmidrule{2-7}
 & Uniform-8 & 40.25 & 44.49 & 47.52 & 49.10 & 43.99 \\
 & Uniform-8 w/o \texttt{FullEvo} & 40.20 & 41.57 & 45.87 & 47.49 & 42.37 \\
\bottomrule
\end{tabular}
\end{table*}

\begin{figure*}[t]
\centering
\includegraphics[width=\textwidth]{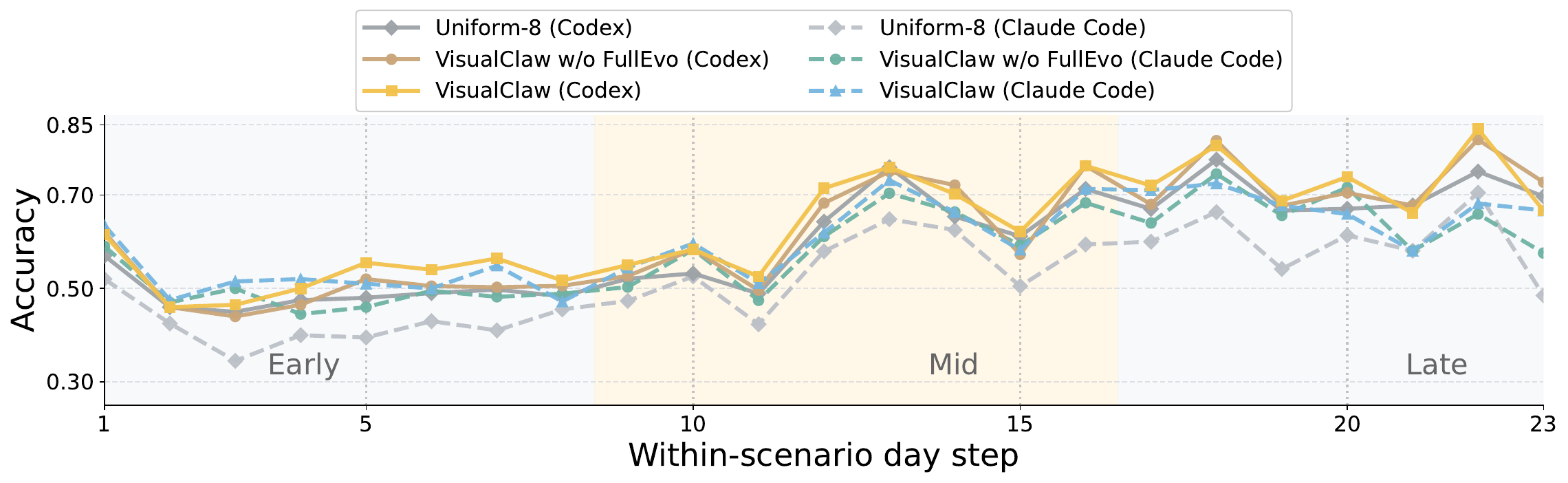}
\caption{Per-day accuracy on 200 scenarios in \mmbench{}. We present \mmmc{} (\texttt{Cat.}), \mmmc{} w/o \texttt{FullEvo}, and Uniform-8. The $x$-axis is within-scenario step; the evolution frequency is set to 5 day steps and only steps with at least $20$ scenarios are displayed.}
\label{fig:mm-arena-per-day}
\end{figure*}

\begin{table}[t]
\centering
\caption{\mmbench{} macro accuracy by empirical scenario difficulty. Tiers are assigned on the same $200$ scenarios using the average Cascade-8 w/o \texttt{FullEvo} accuracy of Codex and Claude Code: Easy $>75\%$, Medium $55$--$75\%$, Hard $\leq55\%$.}
\label{tab:mm-arena-difficulty}
\scriptsize
\setlength{\tabcolsep}{8pt}
\resizebox{.9\columnwidth}{!}{%
\begin{tabular}{llccc}
\toprule
Backend & Setting & Easy (45) & Medium (46) & Hard (109) \\
\midrule
\multirow{5}{*}{Codex}
 & \mmmc{} (\texttt{Cat.}) & \textbf{86.65} & 65.09 & \textbf{36.33} \\
 & \mmmc{} w/o \texttt{FullEvo} & 86.26 & \textbf{65.53} & 30.95 \\
 \cmidrule{2-5}
 & Uniform-8 & 84.29 & 60.42 & 31.90 \\
 & Uniform-8 w/o \texttt{FullEvo} & 85.83 & 58.19 & 29.03 \\
\midrule
\multirow{5}{*}{Claude Code}
 & \mmmc{} (\texttt{Cat.}) & 83.45 & \textbf{64.74} & \textbf{33.94} \\
 & \mmmc{} w/o \texttt{FullEvo} & \textbf{84.96} & 62.14 & 28.61 \\
\cmidrule{2-5}
 & Uniform-8 & 76.60 & 52.41 & 26.97 \\
 & Uniform-8 w/o \texttt{FullEvo} & 77.73 & 49.92 & 24.58 \\
\bottomrule
\end{tabular}
}
\end{table}

\paragraph{Agentic \texttt{FullEvo} Transfers Beyond video-QA.}
Table~\ref{tab:mm-arena} shows consistent gains across both agent backends on the matched $3{,}106$-round core. With Codex, \mmmc{} (\texttt{Cat.}) reaches $54.27\%$ macro accuracy, improving over \mmmc{} w/o \texttt{FullEvo} by $+2.92$ points and Uniform-8 by $+4.02$ points. With Claude Code, the same setting reaches $52.16\%$, improving over w/o \texttt{FullEvo} by $+3.16$ points and Uniform-8 by $+8.17$ points. Thus the self-evolution stack remains useful after moving from single-call video-QA to multi-step multimodal agent execution.

\paragraph{\texttt{FullEvo(Cat.)} Fits Complex Agentic Workflows More.}
Another interesting finding from Table~\ref{tab:mm-arena} is that unlike results on static video-QA tasks, simple memory-skill concatenation is strongest in \mmbench{}. The gain is small but positive for Codex ($+0.38$ over \texttt{Guide}) and clearer for Claude Code ($+1.39$), suggesting that agentic backends can better use long context when they process it through file reads, tool calls, edits, and verification loops. In this setting, extra retrieved failures are less likely to act only as prompt clutter; they can be turned into structured intermediate decisions before the final answer.

\paragraph{Temporal Trends on \mmbench{}.}
The early/mid/late columns and Fig.~\ref{fig:mm-arena-per-day} show a generally increasing within-scenario trend rather than a flat pass rate. Accuracy often drops after the first step because early rounds tend to check the scenario contract, while later rounds require the agent to reconcile video evidence, text artifacts, and its own workspace state. After this adaptation period, \mmmc{} benefits from both accumulated scenario context and self-evolved skills: Codex \mmmc{} (\texttt{Cat.}) is strongest in the mid and late thirds ($56.19\%$, $57.38\%$), while Claude Code \mmmc{} (\texttt{Cat.}) leads all three thirds ($52.03\%$, $50.91\%$, $53.50\%$). This suggests that \texttt{FullEvo} is most useful once the agent has enough local trajectory to turn retrieved failures into concrete decisions, rather than only into longer prompts.

\paragraph{Difficulty Trends on \mmbench{}.}
Table~\ref{tab:mm-arena-difficulty} shows that \texttt{FullEvo}'s benefit is difficulty-dependent. Averaged over Codex and Claude Code, \mmmc{} (\texttt{Cat.}) is slightly lower than w/o \texttt{FullEvo} on Easy scenarios ($85.05\%$ vs.\ $85.61\%$), but improves Medium scenarios ($64.92\%$ vs.\ $63.84\%$) and strongly improves Hard scenarios ($35.14\%$ vs.\ $29.78\%$). The same pattern appears for Uniform-8: adding \texttt{FullEvo} reduces the Easy average ($80.45\%$ vs.\ $81.80\%$) but improves Medium ($56.63\%$ vs.\ $54.60\%$) and Hard ($29.44\%$ vs.\ $26.81\%$). This matches our interpretation: easy tasks are already near saturation, so extra skills and retrieved failures can add prompt noise or over-correct simple answers; medium and hard tasks have more cross-modal conflict, longer dependencies, and file-level constraints, where evolved skills provide useful procedural bias. 
Across sources, Indoor/VSI remains the hardest split ($\sim\!36$--$38\%$ macro under \texttt{Cat.}), EgoSchema is the easiest ($\sim\!81$--$84\%$), and QVHighlights sits in between ($\sim\!55$--$57\%$), so the benchmark is not solved by one source family. We further present more statistics in Appendix~\ref{app:mm-arena}.

\begin{table}[t]
\centering
\caption{Per-benchmark API-cost comparison on Gemini~3~Flash. KF/Q = frames sent to the API; tok/Q = input tokens per question; \$/run = Gemini~3~Flash spend across the benchmark. Savings are our \mmmc{} vs the two baselines.}
\label{tab:cost-headline}
\scriptsize
\setlength{\tabcolsep}{3pt}
\begin{tabular}{llrrrcc}
\toprule
Dataset (average duration) & Configuration & KF/Q & tok/Q & \$/run & vs Full-frame & vs U-8+\texttt{FullEvo} \\
\midrule
\multirow{3}{*}{EgoSchema (3\,min)}
 & Full-frame @1\,fps                          & $\sim$180 & $\sim$192{,}841   & \$28.93   & --- & --- \\
 & Uniform-8 + \texttt{FullEvo}                & 8.00      & 13{,}419          & \$2.01    & --- & --- \\
 & \textbf{\mmmc{}}  & 2.95      & 9{,}524           & \textbf{\$1.44} & \textcolor{blue}{$-95.0$\%} & \textcolor{blue}{$-28.4$\%} \\
\midrule
\multirow{3}{*}{V-MME long ($\sim$30\,min)}
 & Full-frame @1\,fps                          & $\sim$1{,}800 & $\sim$1{,}926{,}361 & \$520.12 & --- & --- \\
 & Uniform-8 + \texttt{FullEvo}                & 8.00          & 15{,}818            & \$4.28   & --- & --- \\
 & \textbf{\mmmc{}}  & 5.41          & 13{,}420            & \textbf{\$3.63} & \textcolor{blue}{$-99.3$\%} & \textcolor{blue}{$-15.2$\%} \\
\midrule
\multirow{3}{*}{EgoPlan-Bench ($\sim$23\,s)}
 & Full-frame @1\,fps                          & $\sim$23 & $\sim$16{,}363  & \$4.53    & --- & --- \\
 & Uniform-8 + \texttt{FullEvo}                & 8.00     & 13{,}348        & \$3.69    & --- & --- \\
 & \textbf{\mmmc{}}  & 1.13     & 10{,}728        & \textbf{\$2.97} & \textcolor{blue}{$-34.4$\%} & \textcolor{blue}{$-19.5$\%} \\
\midrule
\multirow{3}{*}{NextQA ($\sim$30\,s)}
 & Full-frame @1\,fps                          & $\sim$30 & $\sim$32{,}429 & \$9.73    & --- & --- \\
 & Uniform-8 + \texttt{FullEvo}                & 8.00     & 14{,}025       & \$4.21    & --- & --- \\
 & \textbf{\mmmc{}}  & 1.51     & 8{,}207        & \textbf{\$2.47} & \textcolor{blue}{$-74.6$\%} & \textcolor{blue}{$-41.3$\%} \\
\midrule
\multicolumn{2}{l}{\textbf{All experiments total} ($n{=}3{,}322$\,Q, $4$ benchmarks)}
 & --- & --- & \$563.31 / \$14.19 / \textbf{\$10.51} & \textcolor{blue}{$-98.1$\%} & \textcolor{blue}{$-25.9$\%} \\
\bottomrule
\end{tabular}
\end{table}

\subsection{Efficiency Results and Analysis}
\label{sec:efficiency}

Beyond accuracy, the proposed cascade gate plus hybrid injection also cuts API cost materially. We compare frames, input tokens, and the API spend on Gemini~3~Flash against (a)~Full-frame @$1$\,fps and (b)~the Uniform-$8$+\texttt{FullEvo} offline upperbound from Sec.~\ref{sec:headline}. We focus on cost and call-count behavior in this section.

\paragraph{Cascade Cuts API Cost by an Order of Magnitude.}
In Table~\ref{tab:cost-headline}, we compare our cascade-driven video encoding against two baselines on Gemini~3~Flash: the mainstream Full-frame @1\,fps and the offline upperbound Uniform-8 + \texttt{FullEvo}.
Against Full-frame @1\,fps, our cascade ships $2$--$340\times$ fewer frames per question and reduces the cost by $-34.4\%$ to $-99.3\%$ per benchmark, with an average of \textbf{$-98.1\%$} over the four benchmarks and a peak of $-99.3\%$ on V-MME long where $30$-min clips would otherwise consume $\sim\!1.93$M input tokens per question.
Notably, the saving widens with clip duration ($-95.0\%$ on $3$-min EgoSchema, $-99.3\%$ on $30$-min V-MME long, $-34.4\%$ on the single-frame EgoPlan-Bench), evidencing that the cascade's dHash gate is most effective on long-form streaming workloads.
Moreover, against the offline upperbound under the same skill bank configuration, our cascade still ships only $1.13$--$5.41$ frames per question and undercuts the upperbound's cost by $-15.2\%$ to $-41.3\%$ per benchmark (average $-25.9\%$, peak $-41.3\%$ on NextQA), while tracking its accuracy on long clips within $5.20\%$ on EgoSchema and $2.56\%$ on V-MME long.
Overall, the cumulative spend across the four benchmarks drops from \$$563.31$ to $\textbf{\$10.51}$ at \texttt{FullEvo}, and end-to-end latency is dominated by the cloud round-trip rather than the cascade itself ($<\!10$\,ms on-device, $>\!100$\,fps on CPU; full profile in Appendix~\ref{app:full-token-profile}).

\begin{table}[t]
\centering
\caption{Agent/evolver call counts and locally available cost on \mmbench{}.}
\label{tab:mm-arena-claude-cost}
\scriptsize
\setlength{\tabcolsep}{9pt}
\resizebox{.95\columnwidth}{!}{%
\begin{tabular}{llrrrrr}
\toprule
Backend & Setting & Agent Calls & Evolver Calls & Agent Cost & Evolver Cost & Total Cost \\
\midrule
\multirow{3}{*}{Claude Code}
 & Uniform-8 & 3{,}123 & 308 & \$665.68 & \$21.20 & \$686.88 \\
 & Cascade-8 & 3{,}114 & 233 & \$611.51 & \$15.71 & \$627.21 \\
\cmidrule{2-7}
 & $\Delta$ & $-0.3\%$ & $-32\%$ & $-8.9\%$ & $-35.0\%$ & $-9.5\%$ \\
\bottomrule
\end{tabular}
}
\end{table}

\begin{figure*}[t]
\centering
\includegraphics[width=\textwidth]{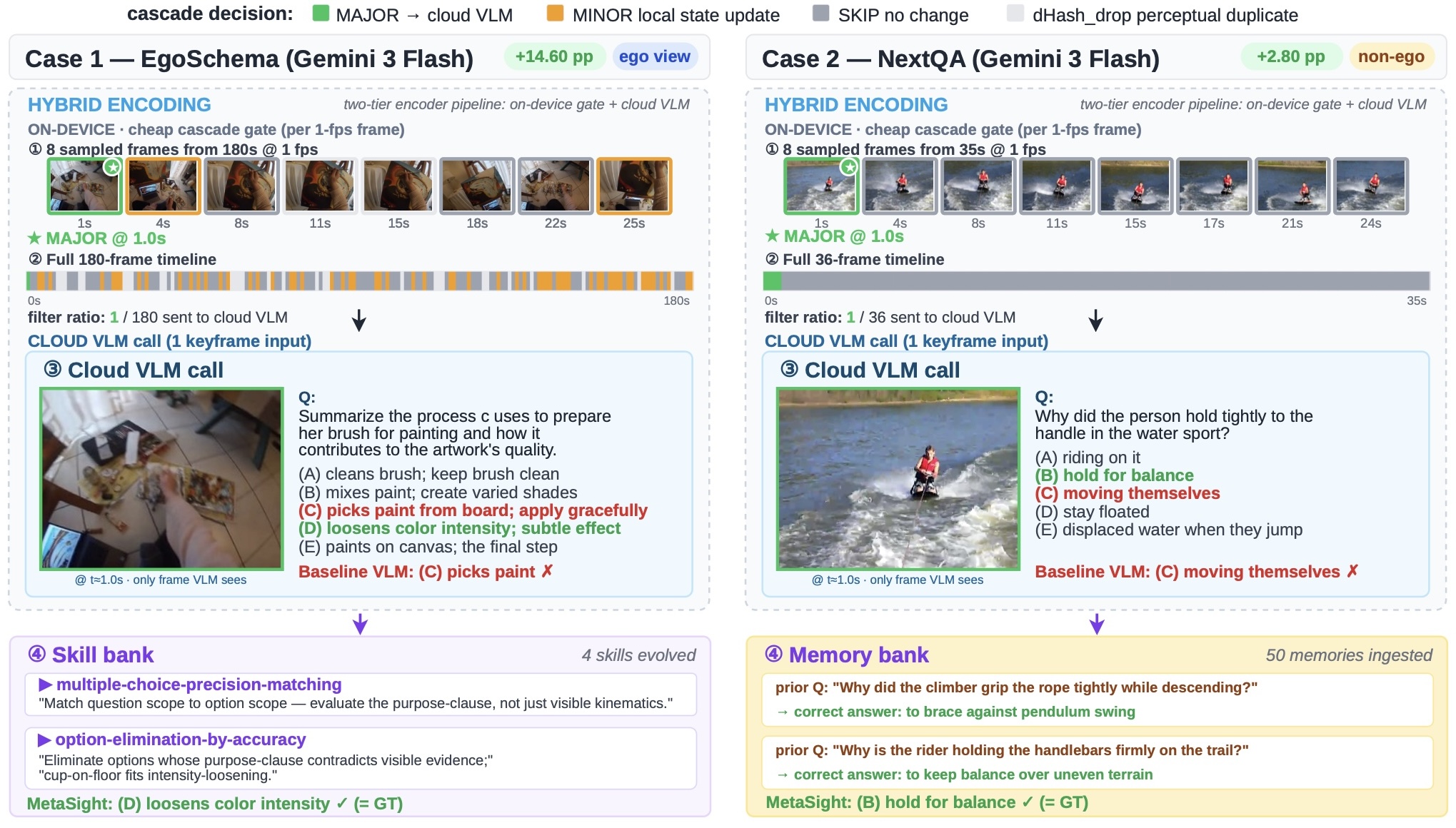}
\caption{
Case studies of \mmmc{}.
Case~1 (EgoSchema, $+\mathbf{14.60\%}$): single \textsc{major} keyframe ($1/180$); evolved skills flip baseline \textbf{(C)} $\to$ GT (D). Case~2 (NextQA, $+\mathbf{2.80\%}$): single keyframe ($1/36$); two memory entries on the ``hold-tight $\to$ stabilise'' flip (C) $\to$ the GT (B)}
\label{fig:case_study_0}
\end{figure*}

\paragraph{Cascade also Wins \wrt Costs on \mmbench{}.}
For the agentic benchmark, we only compare dollar cost for Claude Code because Codex CLI artifacts do not expose provider usage. Table~\ref{tab:mm-arena-claude-cost} shows that Claude Code Uniform-8 is both lower-accuracy and \$$59.67$ more expensive than Cascade-8 under the same v3.1-on setting ($+9.5\%$ total cost). The number of Claude Code agent calls is nearly unchanged ($3{,}123$ vs.\ $3{,}114$), so the cost gap mainly comes from larger agent-side context and more failure-triggered evolution: Uniform-8 spends \$$54.17$ more on agent calls and adds $75$ evolver calls, contributing another \$$5.50$. Codex shows the same call-count direction for the shared evolver component ($146$ vs.\ $120$ calls), but we leave Codex total cost blank. This follows the same efficiency direction as video-QA: cascade selection reduces the visual context seen by the agent, reduces downstream failures, and therefore lowers both the answer-time and evolution-time budgets.

\subsection{Case Studies}
\label{sec:cases}

Figure~\ref{fig:case_study_0} traces two representative \mmmc{} wins on Gemini~3~Flash, one per recovery pathway. Both share the similar efficient encoding signature: the cascade compresses the clip down to a single \textsc{major} keyframe at $t{\approx}1$\,s ($1/180$ and $1/36$), so the swing happens purely at the language layer. 
In Case~1, the baseline locks onto a surface noun on the keyframe (``picks paint'') and skips the question's purpose-clause; the evolved skills inject a more concrete scope discipline at retrieval time and recovers GT (D) ``loosens color intensity.'' Case~2 has no salient skill on retrieval --- the lift instead comes from the memory bank: two prior confidence-gated exemplars on the ``hold-tight $\to$ stabilise'' pattern (rope-while-descending, handlebars-on-trail) re-route the evolver away from (C) ``moving themselves'' to GT (B) ``hold for balance,''. 
Two further wins of our approach, including a cross-VLM transfer to GPT-5.2, are presented in Appendix~\ref{app:more-cases}.

\section{Conclusion}
We presented \mmmc{}, a self-evolving multimodal agent for efficient and adaptive deployment, and \mmbench{}, a curated multimodal agentic benchmark for evaluating visual evidence use inside tool-using workspaces. \mmmc{} is built on hybrid encoding at three timescales: an edge cascade that filters frames before upload, hot/cold skill injection that bounds prompt cost as the bank grows, and memory routed into an offline evolver rather than concatenated per question. Across four video-QA benchmarks and two VLM families, \mmmc{} cuts API cost substantially while improving accuracy in most settings without any weight updates. On \mmbench{}, the same self-evolution stack improves macro accuracy over no-evolution baselines, suggesting the mechanism transfers beyond standard MC video-QA. Together these results address streaming cost, static scaffolds, and missing agentic evaluation for deployable multimodal agents.

{
    \small
    \bibliographystyle{unsrt}
    \bibliography{reference}

@inproceedings{zhang2024llovi,
  title     = {A Simple {LLM} Framework for Long-Range Video Question-Answering},
  author    = {Zhang, Ce and Lu, Taixi and Islam, Md Mohaiminul and Wang, Ziyang and Yu, Shoubin and Bansal, Mohit and Bertasius, Gedas},
  booktitle = {Proceedings of the 2024 Conference on Empirical Methods in Natural Language Processing (EMNLP)},
  year      = {2024},
  note      = {arXiv:2312.17235},
}

@inproceedings{song2024moviechat,
  title     = {{MovieChat}: From Dense Token to Sparse Memory for Long Video Understanding},
  author    = {Song, Enxin and Chai, Wenhao and Wang, Guanhong and Zhang, Yucheng and Zhou, Haoyang and Wu, Feiyang and Chi, Haozhe and Guo, Xun and Ye, Tian and Zhang, Yanting and Lu, Yan and Hwang, Jenq-Neng and Wang, Gaoang},
  booktitle = {Proceedings of the IEEE/CVF Conference on Computer Vision and Pattern Recognition (CVPR)},
  year      = {2024},
  note      = {arXiv:2307.16449},
}

@inproceedings{he2024malmm,
  title     = {{MA-LMM}: Memory-Augmented Large Multimodal Model for Long-Term Video Understanding},
  author    = {He, Bo and Li, Hengduo and Jang, Young Kyun and Jia, Menglin and Cao, Xuefei and Shah, Ashish and Shrivastava, Abhinav and Lim, Ser-Nam},
  booktitle = {Proceedings of the IEEE/CVF Conference on Computer Vision and Pattern Recognition (CVPR)},
  year      = {2024},
  note      = {arXiv:2404.05726},
}

@inproceedings{wang2024videoagent,
  title     = {{VideoAgent}: Long-form Video Understanding with Large Language Model as Agent},
  author    = {Wang, Xiaohan and Zhang, Yuhui and Zohar, Orr and Yeung-Levy, Serena},
  booktitle = {Proceedings of the European Conference on Computer Vision (ECCV)},
  year      = {2024},
  note      = {arXiv:2403.10517},
}

@inproceedings{yu2025framevoyager,
  title     = {{Frame-Voyager}: Learning to Query Frames for Video Large Language Models},
  author    = {Yu, Sicheng and Jin, Chengkai and Wang, Huanyu and Chen, Zhenghao and Jin, Sheng and Zuo, Zhongrong and Xu, Xiaolei and Sun, Zhenbang and Zhang, Bingni and Wu, Jiawei and Zhang, Hao and Sun, Qianru},
  booktitle = {The Thirteenth International Conference on Learning Representations (ICLR)},
  year      = {2025},
  note      = {arXiv:2410.03226},
}

@inproceedings{yu2023sevila,
  title     = {Self-Chained Image-Language Model for Video Localization and Question Answering},
  author    = {Yu, Shoubin and Cho, Jaemin and Yadav, Prateek and Bansal, Mohit},
  booktitle = {Advances in Neural Information Processing Systems (NeurIPS)},
  year      = {2023},
  note      = {arXiv:2305.06988},
}

@inproceedings{wang2025videotree,
  title     = {{VideoTree}: Adaptive Tree-based Video Representation for {LLM} Reasoning on Long Videos},
  author    = {Wang, Ziyang and Yu, Shoubin and Stengel-Eskin, Elias and Yoon, Jaehong and Cheng, Feng and Bertasius, Gedas and Bansal, Mohit},
  booktitle = {Proceedings of the IEEE/CVF Conference on Computer Vision and Pattern Recognition (CVPR)},
  year      = {2025},
  note      = {arXiv:2405.19209},
}

@inproceedings{ren2024timechat,
  title     = {{TimeChat}: A Time-sensitive Multimodal Large Language Model for Long Video Understanding},
  author    = {Ren, Shuhuai and Yao, Linli and Li, Shicheng and Sun, Xu and Hou, Lu},
  booktitle = {Proceedings of the IEEE/CVF Conference on Computer Vision and Pattern Recognition (CVPR)},
  year      = {2024},
  note      = {arXiv:2312.02051},
}

@inproceedings{huang2024vtimellm,
  title     = {{VTimeLLM}: Empower {LLM} to Grasp Video Moments},
  author    = {Huang, Bin and Wang, Xin and Chen, Hong and Song, Zihan and Zhu, Wenwu},
  booktitle = {Proceedings of the IEEE/CVF Conference on Computer Vision and Pattern Recognition (CVPR)},
  year      = {2024},
  note      = {arXiv:2311.18445},
}

@inproceedings{yang2023vid2seq,
  title     = {{Vid2Seq}: Large-Scale Pretraining of a Visual Language Model for Dense Video Captioning},
  author    = {Yang, Antoine and Nagrani, Arsha and Seo, Paul Hongsuck and Miech, Antoine and Pont-Tuset, Jordi and Laptev, Ivan and Sivic, Josef and Schmid, Cordelia},
  booktitle = {Proceedings of the IEEE/CVF Conference on Computer Vision and Pattern Recognition (CVPR)},
  year      = {2023},
  note      = {arXiv:2302.14115},
}

@inproceedings{zhang2023videollama,
  title     = {{Video-LLaMA}: An Instruction-tuned Audio-Visual Language Model for Video Understanding},
  author    = {Zhang, Hang and Li, Xin and Bing, Lidong},
  booktitle = {Proceedings of the 2023 Conference on Empirical Methods in Natural Language Processing: System Demonstrations (EMNLP Demo)},
  year      = {2023},
  note      = {arXiv:2306.02858},
}

@article{zhang2025llavavideo,
  title   = {Video Instruction Tuning With Synthetic Data},
  author  = {Zhang, Yuanhan and Wu, Jinming and Li, Wei and Li, Bo and Ma, Zejun and Liu, Ziwei and Li, Chunyuan},
  journal = {Transactions on Machine Learning Research},
  year    = {2025},
  note    = {arXiv:2410.02713; introduces LLaVA-Video},
}

@inproceedings{papalampidi2024longvivit,
  title     = {A Simple Recipe for Contrastively Pre-training Video-First Encoders Beyond 16 Frames},
  author    = {Papalampidi, Pinelopi and Koppula, Skanda and Pathak, Shreya and Chiu, Justin and Heyward, Joe and Patraucean, Viorica and Shen, Jiajun and Miech, Antoine and Zisserman, Andrew and Nematzadeh, Aida},
  booktitle = {Proceedings of the IEEE/CVF Conference on Computer Vision and Pattern Recognition (CVPR)},
  year      = {2024},
  note      = {arXiv:2312.07395; introduces LongViViT},
}

@inproceedings{shinn2023reflexion,
  title     = {Reflexion: Language Agents with Verbal Reinforcement Learning},
  author    = {Shinn, Noah and Cassano, Federico and Berman, Edward and Gopinath, Ashwin and Narasimhan, Karthik and Yao, Shunyu},
  booktitle = {Advances in Neural Information Processing Systems (NeurIPS)},
  year      = {2023},
  note      = {arXiv:2303.11366},
}

@article{wang2024voyager,
  title   = {Voyager: An Open-Ended Embodied Agent with Large Language Models},
  author  = {Wang, Guanzhi and Xie, Yuqi and Jiang, Yunfan and Mandlekar, Ajay and Xiao, Chaowei and Zhu, Yuke and Fan, Linxi and Anandkumar, Anima},
  journal = {Transactions on Machine Learning Research},
  year    = {2024},
  note    = {arXiv:2305.16291},
}

@inproceedings{zhao2024expel,
  title     = {{ExpeL}: {LLM} Agents Are Experiential Learners},
  author    = {Zhao, Andrew and Huang, Daniel and Xu, Quentin and Lin, Matthieu and Liu, Yong-Jin and Huang, Gao},
  booktitle = {Proceedings of the AAAI Conference on Artificial Intelligence (AAAI)},
  year      = {2024},
  note      = {arXiv:2308.10144},
}

@misc{tang2025agentkb,
  title         = {Agent {KB}: Leveraging Cross-Domain Experience for Agentic Problem Solving},
  author        = {Tang, Xiangru and Qin, Tianrui and Peng, Tianhao and Zhou, Ziyang and Shao, Daniel and Du, Tingting and Wei, Xinming and Xia, Peng and Wu, Fang and Zhu, He and Zhang, Ge and Liu, Jiaheng and Wang, Xingyao and Hong, Sirui and Wu, Chenglin and Cheng, Hao and Wang, Chi and Zhou, Wangchunshu},
  year          = {2025},
  eprint        = {2507.06229},
  archivePrefix = {arXiv},
  primaryClass  = {cs.AI},
}

@misc{packer2023memgpt,
  title         = {{MemGPT}: Towards {LLMs} as Operating Systems},
  author        = {Packer, Charles and Wooders, Sarah and Lin, Kevin and Fang, Vivian and Patil, Shishir G. and Stoica, Ion and Gonzalez, Joseph E.},
  year          = {2023},
  eprint        = {2310.08560},
  archivePrefix = {arXiv},
  primaryClass  = {cs.AI},
}

@inproceedings{park2023generative,
  title     = {Generative Agents: Interactive Simulacra of Human Behavior},
  author    = {Park, Joon Sung and O'Brien, Joseph C. and Cai, Carrie J. and Morris, Meredith Ringel and Liang, Percy and Bernstein, Michael S.},
  booktitle = {Proceedings of the 36th Annual ACM Symposium on User Interface Software and Technology (UIST)},
  year      = {2023},
  note      = {arXiv:2304.03442},
}

@misc{chhikara2025mem0,
  title         = {{Mem0}: Building Production-Ready {AI} Agents with Scalable Long-Term Memory},
  author        = {Chhikara, Prateek and Khant, Dev and Aryan, Saket and Singh, Taranjeet and Yadav, Deshraj},
  year          = {2025},
  eprint        = {2504.19413},
  archivePrefix = {arXiv},
  primaryClass  = {cs.CL},
}

@misc{zhang2025memevolve,
  title         = {{MemEvolve}: Meta-Evolution of Agent Memory Systems},
  author        = {Zhang, Guibin and Ren, Haotian and Zhan, Chong and Zhou, Zhenhong and Wang, Junhao and Zhu, He and Zhou, Wangchunshu and Yan, Shuicheng},
  year          = {2025},
  eprint        = {2512.18746},
  archivePrefix = {arXiv},
  primaryClass  = {cs.AI},
}

@misc{duan2016rl2,
  title         = {{RL$^2$}: Fast Reinforcement Learning via Slow Reinforcement Learning},
  author        = {Duan, Yan and Schulman, John and Chen, Xi and Bartlett, Peter L. and Sutskever, Ilya and Abbeel, Pieter},
  year          = {2016},
  eprint        = {1611.02779},
  archivePrefix = {arXiv},
  primaryClass  = {cs.AI},
}

@inproceedings{rakelly2019pearl,
  title     = {Efficient Off-Policy Meta-Reinforcement Learning via Probabilistic Context Variables},
  author    = {Rakelly, Kate and Zhou, Aurick and Finn, Chelsea and Levine, Sergey and Quillen, Deirdre},
  booktitle = {Proceedings of the 36th International Conference on Machine Learning (ICML)},
  year      = {2019},
  note      = {arXiv:1903.08254},
}

@inproceedings{rothfuss2019promp,
  title     = {{ProMP}: Proximal Meta-Policy Search},
  author    = {Rothfuss, Jonas and Lee, Dennis and Clavera, Ignasi and Asfour, Tamim and Abbeel, Pieter},
  booktitle = {International Conference on Learning Representations (ICLR)},
  year      = {2019},
  note      = {arXiv:1810.06784},
}

@article{kirkpatrick2017ewc,
  title   = {Overcoming Catastrophic Forgetting in Neural Networks},
  author  = {Kirkpatrick, James and Pascanu, Razvan and Rabinowitz, Neil and Veness, Joel and Desjardins, Guillaume and Rusu, Andrei A. and Milan, Kieran and Quan, John and Ramalho, Tiago and Grabska-Barwinska, Agnieszka and Hassabis, Demis and Clopath, Claudia and Kumaran, Dharshan and Hadsell, Raia},
  journal = {Proceedings of the National Academy of Sciences (PNAS)},
  volume  = {114},
  number  = {13},
  pages   = {3521--3526},
  year    = {2017},
  note    = {arXiv:1612.00796},
}

@inproceedings{chaudhry2019agem,
  title     = {Efficient Lifelong Learning with {A-GEM}},
  author    = {Chaudhry, Arslan and Ranzato, Marc'Aurelio and Rohrbach, Marcus and Elhoseiny, Mohamed},
  booktitle = {International Conference on Learning Representations (ICLR)},
  year      = {2019},
  note      = {arXiv:1812.00420},
}

@inproceedings{zenke2017synaptic,
  title     = {Continual Learning Through Synaptic Intelligence},
  author    = {Zenke, Friedemann and Poole, Ben and Ganguli, Surya},
  booktitle = {Proceedings of the 34th International Conference on Machine Learning (ICML)},
  pages     = {3987--3995},
  year      = {2017},
  note      = {arXiv:1703.04200},
}

@inproceedings{mangalam2023egoschema,
  title     = {{EgoSchema}: A Diagnostic Benchmark for Very Long-form Video Language Understanding},
  author    = {Mangalam, Karttikeya and Akshulakov, Raiymbek and Malik, Jitendra},
  booktitle = {Advances in Neural Information Processing Systems (NeurIPS) Datasets and Benchmarks Track},
  year      = {2023},
  note      = {arXiv:2308.09126},
}

@misc{chen2023egoplan,
  title         = {{EgoPlan-Bench}: Benchmarking Multimodal Large Language Models for Human-Level Planning},
  author        = {Chen, Yi and Ge, Yuying and Ge, Yixiao and Ding, Mingyu and Li, Bohao and Wang, Rui and Xu, Ruifeng and Shan, Ying and Liu, Xihui},
  year          = {2023},
  eprint        = {2312.06722},
  archivePrefix = {arXiv},
  primaryClass  = {cs.CV},
}

@inproceedings{grauman2022ego4d,
  title     = {{Ego4D}: Around the World in 3,000 Hours of Egocentric Video},
  author    = {Grauman, Kristen and Westbury, Andrew and Byrne, Eugene and Chavis, Zachary and Furnari, Antonino and Girdhar, Rohit and Hamburger, Jackson and Jiang, Hao and Liu, Miao and Liu, Xingyu and Martin, Miguel and Nagarajan, Tushar and Radosavovic, Ilija and Ramakrishnan, Santhosh Kumar and Ryan, Fiona and Sharma, Jayant and Wray, Michael and Xu, Mengmeng and Xu, Eric Zhongcong and Zhao, Chen and Bansal, Siddhant and Batra, Dhruv and Cartillier, Vincent and Crane, Sean and Do, Tien and Doulaty, Morrie and Erapalli, Akshay and Feichtenhofer, Christoph and Fragomeni, Adriano and Fu, Qichen and Gebreselasie, Abrham and Gonz{\'a}lez, Cristina and Hillis, James and Huang, Xuhua and Huang, Yifei and Jia, Wenqi and Khoo, Weslie and Kol{\'a}{\v{r}}, J{\'a}chym and Kottur, Satwik and Kumar, Anurag and Landini, Federico and Li, Chao and Li, Yanghao and Li, Zhenqiang and Mangalam, Karttikeya and Modhugu, Raghava and Munro, Jonathan and Murrell, Tullie and Nishiyasu, Takumi and Price, Will and Puentes, Paola Ruiz and Ramazanova, Merey and Sari, Leda and Somasundaram, Kiran and Southerland, Audrey and Sugano, Yusuke and Tao, Ruijie and Vo, Minh and Wang, Yuchen and Wu, Xindi and Yagi, Takuma and Zhao, Ziwei and Zhu, Yunyi and Arbel{\'a}ez, Pablo and Crandall, David and Damen, Dima and Farinella, Giovanni Maria and Fuegen, Christian and Ghanem, Bernard and Ithapu, Vamsi Krishna and Jawahar, C. V. and Joo, Hanbyul and Kitani, Kris and Li, Haizhou and Newcombe, Richard and Oliva, Aude and Park, Hyun Soo and Rehg, James M. and Sato, Yoichi and Shi, Jianbo and Shou, Mike Zheng and Torralba, Antonio and Torresani, Lorenzo and Yan, Mingfei and Malik, Jitendra},
  booktitle = {Proceedings of the IEEE/CVF Conference on Computer Vision and Pattern Recognition (CVPR)},
  year      = {2022},
  note      = {arXiv:2110.07058},
}

@inproceedings{sigurdsson2018charadesego,
  title     = {Actor and Observer: Joint Modeling of First and Third-Person Videos},
  author    = {Sigurdsson, Gunnar A. and Gupta, Abhinav and Schmid, Cordelia and Farhadi, Ali and Alahari, Karteek},
  booktitle = {Proceedings of the IEEE Conference on Computer Vision and Pattern Recognition (CVPR)},
  year      = {2018},
  note      = {arXiv:1804.09627; introduces Charades-Ego},
}

@inproceedings{cheng2024egothink,
  title     = {{EgoThink}: Evaluating First-Person Perspective Thinking Capability of Vision-Language Models},
  author    = {Cheng, Sijie and Guo, Zhicheng and Wu, Jingwen and Fang, Kechen and Li, Peng and Liu, Huaping and Liu, Yang},
  booktitle = {Proceedings of the IEEE/CVF Conference on Computer Vision and Pattern Recognition (CVPR)},
  year      = {2024},
  note      = {arXiv:2311.15596},
}

@misc{cheng2024videgothink,
  title         = {{VidEgoThink}: Assessing Egocentric Video Understanding Capabilities for Embodied {AI}},
  author        = {Cheng, Sijie and Fang, Kechen and Yu, Yangyang and Zhou, Sicheng and Li, Bohao and Tian, Ye and Li, Tingguang and Han, Lei and Liu, Yang},
  year          = {2024},
  eprint        = {2410.11623},
  archivePrefix = {arXiv},
  primaryClass  = {cs.CV},
}

@misc{fu2024videomme,
  title         = {{Video-MME}: The First-Ever Comprehensive Evaluation Benchmark of Multi-modal {LLMs} in Video Analysis},
  author        = {Fu, Chaoyou and Dai, Yuhan and Luo, Yongdong and Li, Lei and Ren, Shuhuai and Zhang, Renrui and Wang, Zihan and Zhou, Chenyu and Shen, Yunhang and Zhang, Mengdan and Chen, Peixian and Li, Yanwei and Lin, Shaohui and Zhao, Sirui and Li, Ke and Xu, Tong and Zheng, Xiawu and Chen, Enhong and Shan, Caifeng and He, Ran and Sun, Xing},
  year          = {2024},
  eprint        = {2405.21075},
  archivePrefix = {arXiv},
  primaryClass  = {cs.CV},
}

@inproceedings{xiao2021nextqa,
  title     = {{NExT-QA}: Next Phase of Question-Answering to Explaining Temporal Actions},
  author    = {Xiao, Junbin and Shang, Xindi and Yao, Angela and Chua, Tat-Seng},
  booktitle = {Proceedings of the IEEE/CVF Conference on Computer Vision and Pattern Recognition (CVPR)},
  year      = {2021},
  pages     = {9777--9786},
  note      = {arXiv:2105.08276},
}

@inproceedings{li2023intentqa,
  title     = {{IntentQA}: Context-aware Video Intent Reasoning},
  author    = {Li, Jiapeng and Wei, Ping and Han, Wenjuan and Fan, Lifeng},
  booktitle = {Proceedings of the IEEE/CVF International Conference on Computer Vision (ICCV)},
  year      = {2023},
  pages     = {11963--11974},
}

@misc{google2025gemini3flash,
  title        = {Gemini 3 Flash: Frontier Intelligence Built for Speed},
  author       = {{Google DeepMind}},
  year         = {2025},
  howpublished = {\url{https://deepmind.google/blog/gemini-3-flash-frontier-intelligence-built-for-speed/}},
  note         = {Model release blog post; see also \url{https://blog.google/technology/developers/build-with-gemini-3-flash/}},
}

@misc{geminiteam2024gemini15,
  title         = {Gemini 1.5: Unlocking Multimodal Understanding Across Millions of Tokens of Context},
  author        = {{Gemini Team, Google}},
  year          = {2024},
  eprint        = {2403.05530},
  archivePrefix = {arXiv},
  primaryClass  = {cs.CL},
}

@misc{openai2025gpt52,
  title        = {Update to {GPT-5} System Card: {GPT-5.2}},
  author       = {{OpenAI}},
  year         = {2025},
  howpublished = {\url{https://openai.com/index/gpt-5-system-card-update-gpt-5-2/}},
  note         = {Model card update, December 11, 2025},
}

@misc{anthropic2025haiku45,
  title        = {Introducing {Claude Haiku 4.5}},
  author       = {{Anthropic}},
  year         = {2025},
  howpublished = {\url{https://www.anthropic.com/news/claude-haiku-4-5}},
  note         = {System card: \url{https://anthropic.com/claude-haiku-4-5-system-card}},
}

@inproceedings{reimers2019sentencebert,
  title     = {{Sentence-BERT}: Sentence Embeddings using Siamese {BERT}-Networks},
  author    = {Reimers, Nils and Gurevych, Iryna},
  booktitle = {Proceedings of the 2019 Conference on Empirical Methods in Natural Language Processing and the 9th International Joint Conference on Natural Language Processing (EMNLP-IJCNLP)},
  pages     = {3982--3992},
  year      = {2019},
  note      = {arXiv:1908.10084; underlying paper for the all-MiniLM-L6-v2 sentence-transformers model},
}

@article{xia2026metaclaw,
  title={{MetaClaw}: Just Talk -- An Agent That Meta-Learns and Evolves in the Wild},
  author={Xia, Peng and Chen, Jianwen and Yang, Xinyu and Tu, Haoqin and Liu, Jiaqi and Xiong, Kaiwen and Han, Siwei and Qiu, Shi and Ji, Haonian and Zhou, Yuyin and Zheng, Zeyu and Xie, Cihang and Yao, Huaxiu},
  journal={arXiv preprint arXiv:2603.17187},
  year={2026}
}

@article{liu2026visionclaw,
  title={VisionClaw: Always-On AI Agents through Smart Glasses},
  author={Liu, Xiaoan and Lee, DaeHo and Gonzalez, Eric J and Gonzalez-Franco, Mar and Suzuki, Ryo},
  journal={arXiv preprint arXiv:2604.03486},
  year={2026}
}

@inproceedings{yang2025thinking,
  title={Thinking in space: How multimodal large language models see, remember, and recall spaces},
  author={Yang, Jihan and Yang, Shusheng and Gupta, Anjali W and Han, Rilyn and Fei-Fei, Li and Xie, Saining},
  booktitle={Proceedings of the Computer Vision and Pattern Recognition Conference},
  pages={10632--10643},
  year={2025}
}

@article{lei2021detecting,
  title={Detecting moments and highlights in videos via natural language queries},
  author={Lei, Jie and Berg, Tamara L and Bansal, Mohit},
  journal={Advances in Neural Information Processing Systems},
  volume={34},
  pages={11846--11858},
  year={2021}
}

@article{han2026vlaa,
  title={VLAA-GUI: Knowing When to Stop, Recover, and Search, A Modular Framework for GUI Automation},
  author={Han, Qijun and Tu, Haoqin and Wang, Zijun and Dai, Haoyue and Zhou, Yiyang and Lau, Nancy and Cardenas, Alvaro A and Xu, Yuhui and Xu, Ran and Xiong, Caiming and others},
  journal={arXiv preprint arXiv:2604.21375},
  year={2026}
}

@misc{pointer2026gui,
  title        = {{Pointer Agent}: A new state of the art for computer use},
  author       = {{Pointer}},
  year         = {2026},
  howpublished = {\url{https://www.pointer.ai/blog/sota?utm_source=osworld&utm_medium=leaderboard&utm_campaign=sota}},
}
}

\newpage

\appendix

\section{Appendix}

\subsection{Broader Impact}
\label{app:broader_impact}
\paragraph{Potentials for Edge Device Deployment.}
Streaming wearables are the binding use case: for example, a $1$-hour AI-glasses session at $1$\,fps emits $3{,}600$ frames, which under a frontier-VLM full-frame upload would translate into roughly $3.9$M input tokens and a tail-latency budget incompatible with cellular variance. \mmmc{}'s on-device cascade rejects $\sim\!98\%$ of those frames before any radio is woken, so the same hour ships $5$--$20$ uploads end-to-end --- consistent with the per-question keyframe rates measured in Table~\ref{tab:cost-headline} ($1.13$--$5.41$ KF/Q across the four benchmarks). Combined with \texttt{FullEvo}'s skill-driven accuracy lift (Sec.~\ref{sec:headline}), the cascade therefore makes a class of AI-glasses video-QA agents viable that would otherwise be either too expensive (full-frame) or too inaccurate (uniform-8 plain) to deploy on a mobile data plan.

\paragraph{Failure Modes, Drift, and Dual-Use.}
A self-evolving bank can drift in directions a static system cannot. Two operational risks deserve flagging. \emph{(i)~Reward hacking}: the evolver synthesises skills from automatically-scored failures, so any systematic bias in the answer-key (\eg, position bias on MC, length-prior on free-form) can be amplified into the bank over many evolve rounds; we mitigate this with confidence-gated memory ingest and the F1/F2 hygiene filters (Sec.~\ref{sec:memevolve}, Appendix~\ref{app:bank-hygiene}), but at scale a periodic human audit of the bank diff is recommended. \emph{(ii)~Capability-conditional regressions}: as we surface in Sec.~\ref{sec:capability}, a skill bank evolved against one VLM's failure modes can hurt another VLM family --- the same artefact is helpful or harmful depending on backbone, which is unusual relative to today's static-prompt practice and warrants conservative defaults at deployment. On dual-use: the cascade plus skill bank also lowers the cost of always-on visual surveillance pipelines that we did not design for. The architectural property that makes legitimate wearable assistants viable (cheap streaming inference at the edge) is the same property that makes covert deployments cheaper; we view downstream policy and platform-level access controls --- rather than the gate itself --- as the appropriate locus of governance.

\subsection{Further Analysis}
\label{app:perf-analysis}

In this section, we analyze three properties that decide the deployability: cross-VLM transfer of the evolved bank, the runtime behaviour of the skill bank and memory store, and the hot/cold injection trade-off.

\paragraph{Capability-conditional Gains.}
\label{sec:capability}
The performance boosts of our \mmmc{} largely depends on whether the target VLM exhibits the failure modes the bank was evolved to address, not on raw capability. \texttt{FullEvo} delivers $+15.80\%$ on the weaker Gemini~3~Flash but only $+4.00\%$ using the stronger GPT-5.2 on EgoSchema, despite GPT-5.2 starting from an $11.40\%$ higher Plain baseline ($64.00$ vs.\ $52.60$). 
This suggests that GPT-5.2, as a stronger VLM, already possesses much of the commonsense and task knowledge that the evolved skills and memory aim to inject. Nevertheless, \mmmc{} still provides a clear improvement on top of this strong baseline even with the same initial skill bank as Gemini's, verifying that our evolved framework remains useful even when the backbone VLM is already highly capable.
We further provide per-skill transfer dynamics in Appendix~\ref{app:xfer} for better understanding of our method.

\begin{figure}[t]
\begin{minipage}[t]{0.48\linewidth}
\makeatletter\def\@captype{table}
\setlength\tabcolsep{3pt}
\caption{Skill bank composition at end-of-run, Gemini~3~Flash with \texttt{FullEvo}. Total $= K_{\text{seed}}$ initial seed (kept) $+$ evolved (added by the evolver).}
\label{tab:bank-composition}
\scriptsize
\centering
\resizebox{.9\textwidth}{!}{\begin{tabular}{lccc}
\toprule
Bench & Seed & Evolved & Total \\
\midrule
EgoSchema       & 12 & 28 & 40 \\
Video-MME long  & 12 & 45 & 57 \\
EgoPlan-Bench         & 12 & 55 & 67 \\
NextQA          & 12 & 32 & 44 \\
\bottomrule
\end{tabular}
}
\end{minipage}
\hspace{.8pt}
\begin{minipage}[t]{0.48\linewidth}
\centering
\makeatletter\def\@captype{table}
\setlength\tabcolsep{2.5pt}
\caption{Memory ingest and retrieval rates per benchmark using \texttt{FullEvo} and Gemini 3 Flash. Retrievals are skill evolver-fusion fetches, where memory was never injected into VLMs.}
\label{tab:memory-retrieval}
\scriptsize
\resizebox{\textwidth}{!}{\begin{tabular}{lccc}
\toprule
Bench & Ingests & Retrievals & Retrieval/Q \\
\midrule
EgoSchema       & 340 & 11 & 0.022 \\
Video-MME long  & 578 & 22 & 0.024 \\
EgoPlan-Bench         & 266 & 44 & 0.048 \\
NextQA          & 745 & 17 & 0.017 \\
\bottomrule
\end{tabular}
}
\end{minipage}
\vspace{-8pt}
\end{figure}

\paragraph{Bank Composition and Memory Dynamics.}
\label{sec:bank-mem-analysis}
The evolved bank grows substantially per run ($28$--$55$ new skills on a 12-seed base; Table~\ref{tab:bank-composition}), while memory retrieval fires only $\sim\!1\times$ per $21$--$59$ questions (Table~\ref{tab:memory-retrieval}) --- a low-frequency, high-precision conditioning signal for the evolver. F1 ran on every evolve but logged zero rejections at our $0.5$ Jaccard threshold (the Haiku evolver's name diversity stays below the cutoff at this scale); F2 is opt-in and was kept off in the headline grid to keep the ablation uniform, with activation reported on supplementary benchmarks in Appendix~\ref{app:bank-hygiene}. Both are bank-hygiene insurance for longer evolution histories rather than active levers here. On V-MME long the $12$ seed skills accumulate $219$ activations at $73\%$ while the top-$3$ evolved skills accumulate $314$ activations at $71\%$, complementing the findings in Sec.~\ref{sec:headline} that the seed skill bank carries most of the lift on long-form video data (see Appendix~\ref{app:per-skill} for score details). The cosine gate filters memory ingest in step with plain accuracy (NextQA $74.5\%$ down to EgoPlan-Bench $28.8\%$), and retrieval fires only when the skill evolver runs, so the evolver receives $\sim\!40\times$ less memory traffic than a per-question concatenation would deliver.

\paragraph{Hybrid Injection Top-$k$ Trade-off.}
Table~\ref{tab:hybrid-topk} summarizes the hot/cold injection grid as model-wise averages over the four video-QA benchmarks while holding the cascade frame budget and \texttt{FullEvo} stack fixed. Here $K{=}8$ is the fixed frame budget; top-$k$ controls how many hot skills are injected, and \textsc{All} injects the whole skill bank. The preference is backbone-dependent: GPT-5.2 is best at $k{=}3$ ($51.90\%$) and drops under \textsc{All} ($50.12\%$), suggesting that stronger VLMs are more sensitive to unrelated bank entries; Gemini~3~Flash benefits from \textsc{All} ($59.43\%$), consistent with the bank being evolved around Gemini-style failure modes. We therefore use $k{=}3$--$5$ as the robust default when the bank is large or transferred across backbones, and reserve full-bank injection for compact, source-aligned banks or weaker backbones that benefit from more explicit skills.

\begin{table}[t]
\centering
\caption{Model-wise average hybrid hot/cold top-$k$ ablation with \texttt{FullEvo} across the four video-QA benchmarks. The cascade frame budget is fixed at $K{=}8$ for every column; \textsc{All} injects the whole skill bank rather than changing the frame budget. Best value in each row is bolded.}
\label{tab:hybrid-topk}
\scriptsize
\setlength{\tabcolsep}{7pt}
\resizebox{.6\columnwidth}{!}{%
\begin{tabular}{lcccc}
\toprule
Model & $k{=}1$ & $k{=}3$ & $k{=}5$ & \textsc{All} \\
\midrule
GPT-5.2 & 50.48 & \textbf{51.90} & 51.47 & 50.12 \\
Gemini 3 Flash & 57.27 & 57.73 & 57.49 & \textbf{59.43} \\
\bottomrule
\end{tabular}
}
\end{table}

\subsection{Detailed Algorithm}
\label{app:detail-algo}
We detail the algorithm of our \mmmc{} in Algorithm~\ref{alg:streaming} as follow. Our \mmmc{} agentic framework generally consists of a 3-tier gate for hybrid visual encoding and meta-evolution.

\begin{algorithm}[h]
\caption{\mmmc{} agentic streaming inference loop.}
\label{alg:streaming}
\begin{algorithmic}[1]
\Require Meta-model $M=(\theta,S,M_v,G)$; question stream $\{q_i\}$; frame stream $\{f_t\}$; evolve threshold $N_{\text{evo}}$; hybrid top-$k$.
\State $K\gets\emptyset$;\ \ $H\gets\emptyset$;\ \ $D^{\text{fail}}\gets\emptyset$
\For{each frame $f_t$ from edge} \Comment{per-frame, sub-ms}
  \State $h\gets\mathrm{dHash}(f_t)$;\ \ \textbf{continue} if $\min_{h'\in H}\mathrm{Hamming}(h,h')\leq 6$ \Comment{near-dup}
  \State $H\gets H\cup\{h\}$;\ \ $e\gets\mathrm{LightweightEncoder}(f_t)$;\ \ $g\gets\mathrm{ChangeGate}(e;\tau_{\text{major}},\text{decay})$
  \State $K\gets K\cup\{f_t\}$ if $g=\textsc{major}$
\EndFor
\For{each question $q_i$} \Comment{per-question, $\sim\!10$\,ms}
  \State $S_i^{\text{hot}}\gets\mathrm{Ret}_S(q_i,k)$;\ \ $S_i^{\text{cold}}\gets\mathrm{Cat}(S\setminus S_i^{\text{hot}})$;\ \ $m_i\gets\mathrm{Ret}_{M_v}(q_i)$ \Comment{cosine $\geq 0.55$}
  \State $a_i\gets\pi_\theta(\cdot\mid q_i,K,S_i^{\text{hot}},S_i^{\text{cold}})$;\ \ $r_i\gets\mathrm{Score}(a_i,q_i)$;\ \ $\mathrm{UpdateUtility}(S_i^{\text{hot}},r_i)$
  \State $D^{\text{fail}}\gets D^{\text{fail}}\cup\{(q_i,a_i,m_i)\}$ if $r_i=0$
  \If{$|D^{\text{fail}}|\geq N_{\text{evo}}$} \Comment{per-session, minutes}
    \State $\Delta S\gets\mathrm{F1Dedup}(\mathcal{E}(S,D^{\text{fail}},m_i),S)$;\ \ $S\gets S\cup\Delta S$;\ \ $D^{\text{fail}}\gets\emptyset$ \Comment{Jaccard $\geq 0.5$ rejected}
  \EndIf
  \State $S\gets\mathrm{F2Prune}(S)$ every $100$ questions
\EndFor
\end{algorithmic}
\end{algorithm}

\subsection{Compute cost}
\label{app:cost}

Total experiments reported in this paper: $\sim\!\$190$ across Gemini~3~Flash and GPT-5.2 (Azure proxy). Reasoning-token billing on the proxy is not separately exposed and may inflate the GPT-5.2 portion by 1.5--3$\times$ depending on Azure deployment configuration.

\begin{table}[h]
\centering
\caption{Approximate compute costs by stage. ``Wall'' = approximate
wall-clock with 6--12 concurrent runs. All Bedrock Haiku evolver calls are bundled into the Gemini-side accounting.}
\label{tab:cost}
\scriptsize
\begin{tabular}{lccc}
\toprule
Stage & Runs & Approx.\ cost (USD) & Approx.\ wall (h) \\
\midrule
Tier-1 ablation ladder (EgoSchema/V-MME short/TeleEgo) & 16 & 46 & 18 \\
Tier-2 cascade per-stage breakdown & 5 & 2 & 1 \\
Cross-VLM full benchmark (GPT-5.2 6 runs) & 6 & 36 & 8 \\
Headline 4-benchmark $\times$ 2-VLM experiments & 8 & $\sim\!50$ & $\sim\!12$ \\
NextQA + EgoPlan-Bench ablation ladders & 14 & $\sim\!50$ & $\sim\!10$ \\
\midrule
\textbf{Total} & $\sim\!50$ & $\sim\!\mathbf{190}$ & $\sim\!50$ \\
\bottomrule
\end{tabular}
\end{table}

\subsection{Additional \mmbench{} details}
\label{app:mm-arena}

We keep only residual scope and ablation details here; the benchmark construction, leakage gate, and source selection are summarized in Sec.~\ref{sec:mmbench}.

\paragraph{Scope caveat.}
All four core cells are complete and clean for the $200$ active scenarios. The strict exact coverage of the current broadened \texttt{video\_required} selector is $100/200$ scenarios (the EgoSchema and QVHighlights halves); the Indoor/VSI cells still use the earlier narrower \texttt{visual\_required} subset. We therefore report the matched core scope in the main paper and keep broadened-selector runs separate below.

\begin{table}[h]
\centering
\caption{Memory/evolver-context ablations on the matched $3{,}106$-round core scope. ConcatMem simply concatenates retrieved memory into the skill-evolution context; related-prefix inserts it as a guided prefix.}
\label{tab:mm-arena-ablations}
\scriptsize
\setlength{\tabcolsep}{4pt}
\begin{tabular}{llrrr}
\toprule
Backend & Setting & Passed / Total & Micro Acc. & Macro Acc. \\
\midrule
Codex & ConcatMem & 1{,}860 / 3{,}106 & 59.88 & \textbf{54.27} \\
Codex & related-prefix & 1{,}851 / 3{,}106 & 59.59 & 54.02 \\
Codex & assembled FullEvo & 1{,}848 / 3{,}106 & 59.50 & 53.89 \\
Claude Code & ConcatMem & 1{,}795 / 3{,}106 & 57.79 & \textbf{52.16} \\
Claude Code & assembled FullEvo & 1{,}756 / 3{,}106 & 56.54 & 50.77 \\
\bottomrule
\end{tabular}
\end{table}

\paragraph{Ablations and run health.}
ConcatMem improves over assembled FullEvo by $+0.38$ macro points for Codex and $+1.39$ for Claude Code, suggesting that longer raw memory context is more useful in agentic workflows than in one-shot answer selection. Under the broader \texttt{video\_required} selector, the Codex ConcatMem and related-prefix rows both pass $2{,}376/3{,}691$ rounds with $63.02\%$ and $62.98\%$ macro accuracy, respectively, but those rows use a different denominator and are not mixed with Table~\ref{tab:mm-arena}. All reported Codex core/ablation cells have $200/200$ clean exits; Claude Code uses latest clean per-scenario results after rejecting timeout, max-turn, traceback, quota, agent-error, and evolver-failure logs.

\subsection{Cascade-fill on long-clip and uniform-activity benchmarks}
\label{app:cascade-fill}

Table~\ref{tab:cascade-fill-headline} reports parity-budget ($K=8$) cascade-fill on the two short-clip benchmarks where the hybrid wins; we list here the same comparison on EgoSchema (3-min clips, uniform activity) and Video-MME long (30+\,min clips). Per-benchmark \texttt{min\_gap\_s}: EgoSchema $5$\,s, V-MME long $60$\,s, EgoPlan-Bench $1$\,s, NextQA $1$\,s; all runs use \texttt{max-keyframes=8}. GPT-5.2 cascade-fill experiments are not yet measured.

The benchmark-conditional split reflects the underlying frame-relevance distribution. On EgoSchema's 3-min uniform-activity ego clips, salient content is distributed roughly evenly through the clip rather than concentrated in scene transitions, so the cascade gate's content-aware selection has less leverage and uniform-8's even temporal coverage is harder to beat. On V-MME long's 30+\,min clips the situation is different but converges to the same conclusion: with \texttt{min\_gap\_s} forced to $60$\,s to keep the cascade-fill samples non-overlapping, the fill positions are too coarse to recover the content the cascade missed in the long static stretches between scene transitions. We report cascade-fill as a parity-budget alternative that is benchmark-conditional in its win, not a universal replacement for uniform-8.

\subsection{Cross-VLM transfer dynamics}
\label{app:xfer}

\emph{Cross-VLM transfer dynamics are per-skill rather than bank-wide: format-enforcement skills are VLM-family-bound while reasoning-pattern skills transfer cleanly, and the memory-injection mode preference can invert sign across VLM families.} A targeted ablation on the GPT-5.2 EgoSchema row of Table~\ref{tab:main} surfaces three findings.

\paragraph{(i)~Format-enforcement is VLM-family-bound.} Removing the single \texttt{answer-format-completion} skill from the seed bank (\texttt{seed-11}) recovers GPT-5.2 by $+4$ to $+6$\,\% uniformly across three benchmarks at $n{=}50$, because the skill was evolved against a Gemini failure mode (premature abandonment) that GPT-5.2 does not exhibit. \emph{Caveat:} promoting \texttt{seed-11} as a universal bank fails for Gemini at full benchmark ($-4.2$\,\% regression on EgoSchema 500), so per-VLM bank composition is necessary.

\paragraph{(ii)~Reasoning-pattern skills transfer cleanly.} The remaining $11$ reasoning-pattern skills deliver $+2.60$\,\% on GPT-5.2 EgoSchema even without the format skill, mirroring the Gemini ``seed alone is most of the lift'' pattern.

\paragraph{(iii)~Memory-mode preference inverts.} \texttt{+SkillMemCat} costs $-2.0$\,\% on Gemini EgoSchema but delivers $+3.20$\,\% on GPT-5.2 EgoSchema, while memory$\to$evolver fusion is positive on both VLMs --- so direct concatenation is benchmark/VLM-conditional rather than universally hurtful, and we keep \texttt{FullEvo} as the universal recommendation. Combined with the V-MME long picture (full-inject regresses on GPT-5.2; hybrid-3 recovers), the rule is: method transfers cleanly on the evolution-source benchmark, while non-source benchmarks require per-VLM injection-mode tuning.

\subsection{Bank hygiene activity on supplementary benchmark}
\label{app:bank-hygiene}

F1 (token-Jaccard dedup at evolve-time) ran on every evolve but logged zero rejections on the four headline benchmark: at $500$--$1000$ questions per run the Haiku evolver's name diversity stayed below our $0.5$ Jaccard cutoff, so no candidate was rejected as a near-duplicate. F2 (per-skill utility prune) is opt-in (\texttt{--skill-utility-prune}) and was disabled in the headline grid to keep the ablation uniform; on the longer / more diverse-task supplementary benchmark (V-MME short, TeleEgo) where we enabled it, F1 logged $11$ and $5$ rejections respectively, F2 fired $2$ prune events ($10$ skills dropped), delivering a $24$--$37$\,\% bank-size reduction at $\pm\,1$\,\% accuracy. F1+F2 are therefore best understood as low-cost insurance for longer evolution histories rather than active levers in the headline configuration.

\subsection{Per-skill EgoPlan-Bench numerics}
\label{app:per-skill}

On EgoPlan-Bench, all top-tier seed skills bottom out near $27$\,\% accuracy ($225$ activations each), mirroring the benchmark's near-random absolute ceiling for non-frontier VLMs that lack visible task-progress context (Sec.~\ref{sec:capability}). Top evolved skills (\texttt{current-state-action-matching}, \texttt{goal-aligned-action-sequencing}) accumulate $204$ activations at $26$\,\%; later-evolved entries accumulate $148$--$188$ activations at $28$--$29$\,\%. The per-skill spread is below the ablation noise floor on this benchmark, which is consistent with the headline finding that EgoPlan-Bench is at near-random absolute accuracy regardless of method.

\subsection{Cascade per-stage breakdown}
\label{app:cascade-stages}

We retain the per-stage breakdown for transparency, despite production \texttt{cg-adaptive} losing offline accuracy on EgoSchema. Table~\ref{tab:cascade-stages} (plain Gemini~3~Flash, EgoSchema~200, seed=42, $\tau_{\text{major}}=0.20$) shows three things: (i)~uniform-8 wins offline by $+9.5$\,\% at $8$\,KF/Q on this benchmark; (ii)~\texttt{dhash-only} is the worst cascade mode because hash-passed frames cluster toward early video, so deduplication alone without scene-aware reweighting underperforms even uniform sampling; (iii)~adaptive thresholds and temporal decay cost $\sim\!4$\,\% relative to static thresholds at matched KF/Q on EgoSchema. The adaptive features are the only mode robust to live-streaming conditions (slow-moving scenes, stationary cameras, irregular frame arrival), and the cost reverses on diverse-scene benchmark: \texttt{cg-adaptive} wins \texttt{cg-static} by $+3.4$\,\% on TeleEgo. We default to \texttt{cg-adaptive} because cross-benchmark portability under streaming conditions outweighs the EgoSchema-specific offline gap.

\begin{table}[h]
\centering
\caption{Cascade per-stage breakdown on EgoSchema~$200$, plain Gemini~3~Flash, seed=42, $\tau_{\text{major}}=0.20$. KF/Q = avg keyframes per question sent to the VLM.}
\label{tab:cascade-stages}
\small
\begin{tabular}{llccc}
\toprule
Mode & Pipeline & Acc.\,(\%) & KF/Q & Input tokens \\
\midrule
\texttt{uniform-8}    & uniform sampling, no gate                & \textbf{66.0} & 8.00 & 1{,}761{,}645 \\
\texttt{dhash-only}   & dHash $\to$ take all hash-passed         & 57.0 & 8.00 & 1{,}761{,}645 \\
\texttt{cg-static}    & dHash + LE + CG (no decay)               & 60.5 & 4.83 & 1{,}077{,}301 \\
\texttt{cg-adaptive}  & full pipeline (production)               & 56.5 & 4.83 & 1{,}083{,}269 \\
\bottomrule
\end{tabular}
\end{table}

\subsection{Full per-experiment token / cost / latency profile}
\label{app:full-token-profile}

The headline saving in Sec.~\ref{sec:efficiency} (Table~\ref{tab:cost-headline}) compares Gemini Cascade + \texttt{FullEvo} against the Uniform-8 + \texttt{FullEvo} offline ceiling. Table~\ref{tab:tokens} reports the full per-experiment profile for both VLM families across the four benchmarks plus the cross-VLM ablation runs, drawn from the \texttt{vlm\_usage} field of each results dump.

\begin{table}[h]
\centering
\caption{Per-experiment token / cost / latency profile 4 benchmarks $\times$ 2-VLM. \$/Q is the average money (USD) spent per question. lat./Q(s) is the average latency (second) per question on each dataset. KF/Q $=$ avg keyframes per question after the cascade.}
\label{tab:tokens}
\footnotesize
\setlength{\tabcolsep}{4pt}
\begin{tabular}{llcccrrrcc}
\toprule
Bench & Variant & Acc.\,(\%) & in\_tok/Q & out\_tok/Q & \$/run & \$/Q & lat./Q\,(s) & KF/Q \\
\midrule
\multirow{4}{*}{EgoSchema}
 & Gemini plain                & 52.60 & 3{,}430  & 15.1 & \$0.53 & \$0.0011 & 9.05  & 2.96 \\
 & Gemini \texttt{FullEvo}     & 68.00 & 9{,}524  &  5.9 & \$1.44 & \$0.0029 & 9.80  & 2.95 \\
 & GPT-5.2 plain               & 64.00 &    905   &  5.0 & \$0.14 & \$0.0003 & 4.39  & 2.95 \\
 & GPT-5.2 \texttt{FullEvo}    & 68.00 & 6{,}369  &  4.9 & \$0.96 & \$0.0019 & 5.43  & 2.95 \\
\midrule
\multirow{4}{*}{V-MME long}
 & Gemini plain                & 60.33 & 6{,}111  & 10.8 & \$1.67 & \$0.0019 & 14.56 & 5.43 \\
 & Gemini \texttt{FullEvo}     & 64.22 & 13{,}420 &  2.3 & \$3.63 & \$0.0040 & 13.66 & 5.41 \\
 & GPT-5.2 plain               & 55.89 & 5{,}461  &  4.9 & \$6.18 & \$0.0069 & 9.82  & 5.40 \\
 & GPT-5.2 \texttt{FullEvo}    & 55.89 & 6{,}307  &  3.8 & \$7.13 & \$0.0079 & 12.58 & 5.40 \\
\midrule
\multirow{4}{*}{EgoPlan-Bench}
 & Gemini plain                & 24.62 & 1{,}352  & 10.5 & \$0.40 & \$0.0004 & 8.55 & 1.13 \\
 & Gemini \texttt{FullEvo}     & 28.85 & 10{,}728 &  1.5 & \$2.97 & \$0.0032 & 8.99 & 1.13 \\
 & GPT-5.2 plain               & 28.42 & 1{,}198  &  5.1 & \$0.34 & \$0.0004 & 5.43 & 1.13 \\
 & GPT-5.2 \texttt{FullEvo}    & 28.85 & 10{,}207 &  5.2 & \$2.83 & \$0.0031 & 4.11 & 1.13 \\
\midrule
\multirow{4}{*}{NextQA}
 & Gemini plain                & 72.70 & 1{,}750  &  4.0 & \$0.53 & \$0.0005 & 5.70 & 1.51 \\
 & Gemini \texttt{FullEvo}     & 74.50 & 8{,}207  &  1.3 & \$2.47 & \$0.0025 & 5.84 & 1.51 \\
 & GPT-5.2 plain               & 73.20 &    576   &  5.1 & \$0.18 & \$0.0002 & 2.95 & 1.51 \\
 & GPT-5.2 hybrid-3            & 68.10 & 2{,}244  &  4.8 & \$0.68 & \$0.0007 & 9.24 & 1.51 \\
\bottomrule
\end{tabular}
\end{table}

\subsection{Additional case studies}
\label{app:more-cases}
\begin{figure*}[h]
\centering
\includegraphics[width=.95\textwidth]{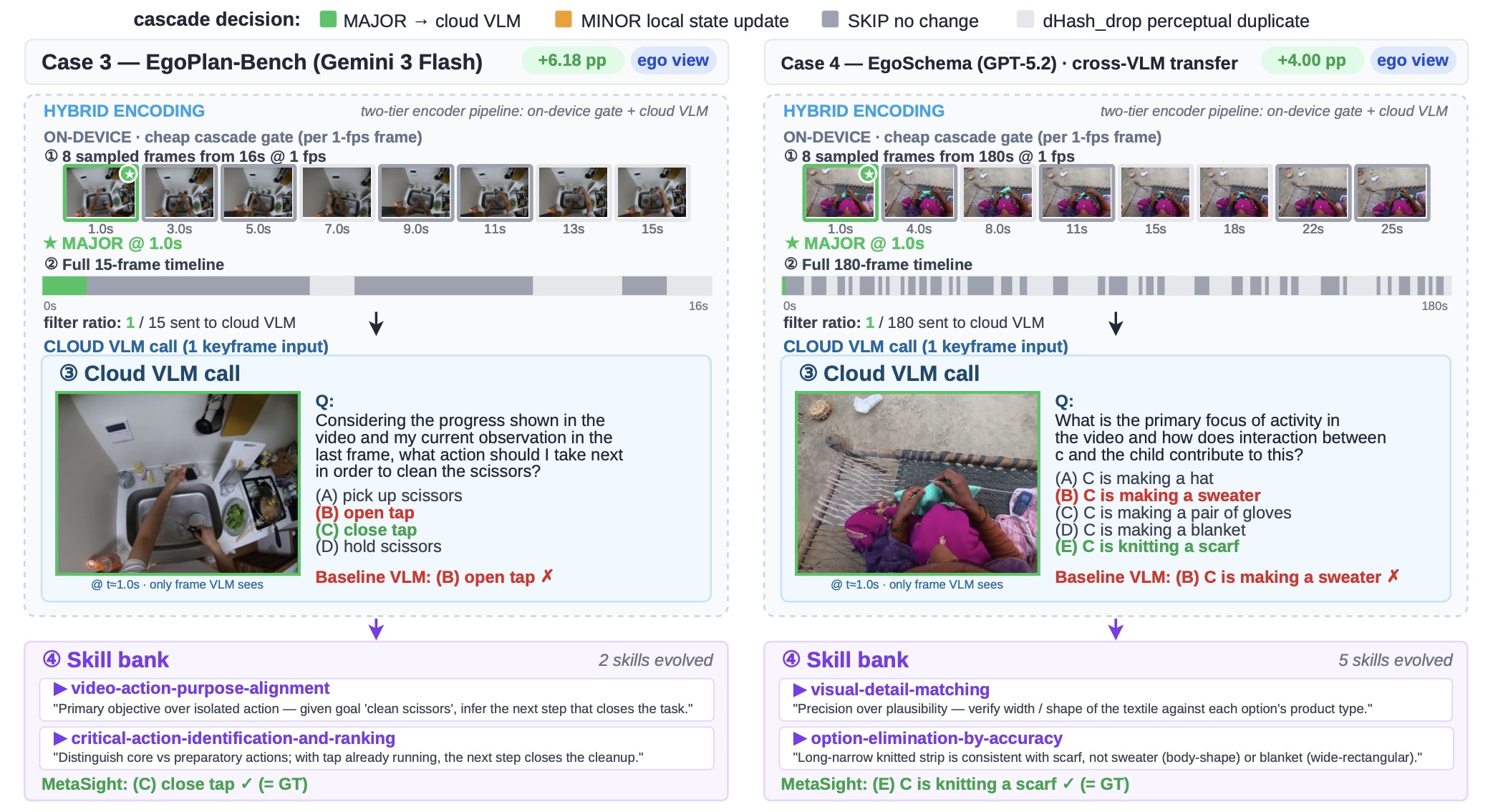}
\caption{Two further \mmmc{} wins, including a cross-VLM transfer. \textbf{Case~3} (EgoPlan-Bench, Gemini~3~Flash, $+\mathbf{6.18\%}$): single keyframe ($1/15$); evolved skills flip \textbf{(B)} ``open tap'' $\to$ GT \textbf{(C)} ``close tap.'' \textbf{Case~4} (EgoSchema, GPT-5.2, $+\mathbf{4.00\%}$, cross-VLM): a Gemini-evolved bank applied unmodified to GPT-5.2 corrects \textbf{(B)} ``making a sweater'' $\to$ GT \textbf{(E)} ``knitting a scarf.''}
\label{fig:case_study_appendix}
\end{figure*}

We provide two more examples of Gemini 3 Flash as well as GPT-5.2 on the EgoPlan-Bench and the EgoSchema benchmark  in Figure~\ref{fig:case_study_appendix}.

\end{document}